\documentclass[10pt,twocolumn,letterpaper]{article}

\usepackage{cvpr}              %

\usepackage[table]{xcolor}
\usepackage[utf8]{inputenc} %
\usepackage[T1]{fontenc}    %
\usepackage{url}            %
\usepackage{booktabs}       %
\usepackage{amsfonts}       %
\usepackage{nicefrac}       %
\usepackage{microtype}      %
\usepackage{amsmath}
\usepackage{multirow}
\usepackage{subcaption}
\usepackage{graphicx}
\usepackage{algorithm}
\usepackage{algorithmic}
\usepackage{wrapfig}
\usepackage{color}
\usepackage{tcolorbox}
\usepackage{nicematrix}
\usepackage{enumitem}
\setlist{leftmargin=8mm}
\usepackage{xfrac}

\renewcommand{\paragraph}[1]{\noindent \textbf{#1}}
\usepackage{lipsum}

\definecolor{cvprblue}{rgb}{0.21,0.49,0.74}
\definecolor{lightgreen}{rgb}{0.,0.3,0.}
\definecolor{stronggreen}{rgb}{0.,0.8,0.}
\definecolor{urlpink}{rgb}{1,0.4,0.5}

\usepackage[pagebackref,breaklinks,colorlinks,hidelinks,allcolors=cvprblue]{hyperref}

\usepackage{pifont}
\usepackage{xcolor}

\def\paperID{16753} %
\def\confName{CVPR}
\def\confYear{2025}

\newcommand{\ourmethod}{\textsc{Switti}\xspace}
\newcommand{\ourmethodar}{\textsc{Switti (AR)}\xspace}

\usepackage{lipsum}
\title{\vspace{-4mm} \ourmethod: Designing Scale-Wise Transformers for Text-to-Image Synthesis}

\author{%
  Anton Voronov$^{1,2,3}$
  \and
  Denis Kuznedelev$^{1,4}$
  \and
  Mikhail Khoroshikh$^{6}$
  \and
  Valentin Khrulkov$^{1,5}$
  \and 
  Dmitry Baranchuk$^{1}$
}

\begin{document}

\twocolumn[{%
\renewcommand\twocolumn[1][]{#1}%
\maketitle
{
\vspace{-8mm}
\hspace{2.75mm}
\footnotemark[1]Yandex Research\hspace{14.25mm}
\footnotemark[2]HSE University\hspace{14.25mm}
\footnotemark[3]MIPT\hspace{14.25mm}
\footnotemark[4]Skoltech\hspace{14.25mm}
\footnotemark[5]AIRI\hspace{14.25mm}
\footnotemark[6]ITMO\vspace{-3mm} \\

\hspace{45mm}\color{urlpink}\small{\url{https://yandex-research.github.io/switti}}
}

\begin{center}
    \centering
    \captionsetup{type=figure}
    \includegraphics[width=\linewidth]{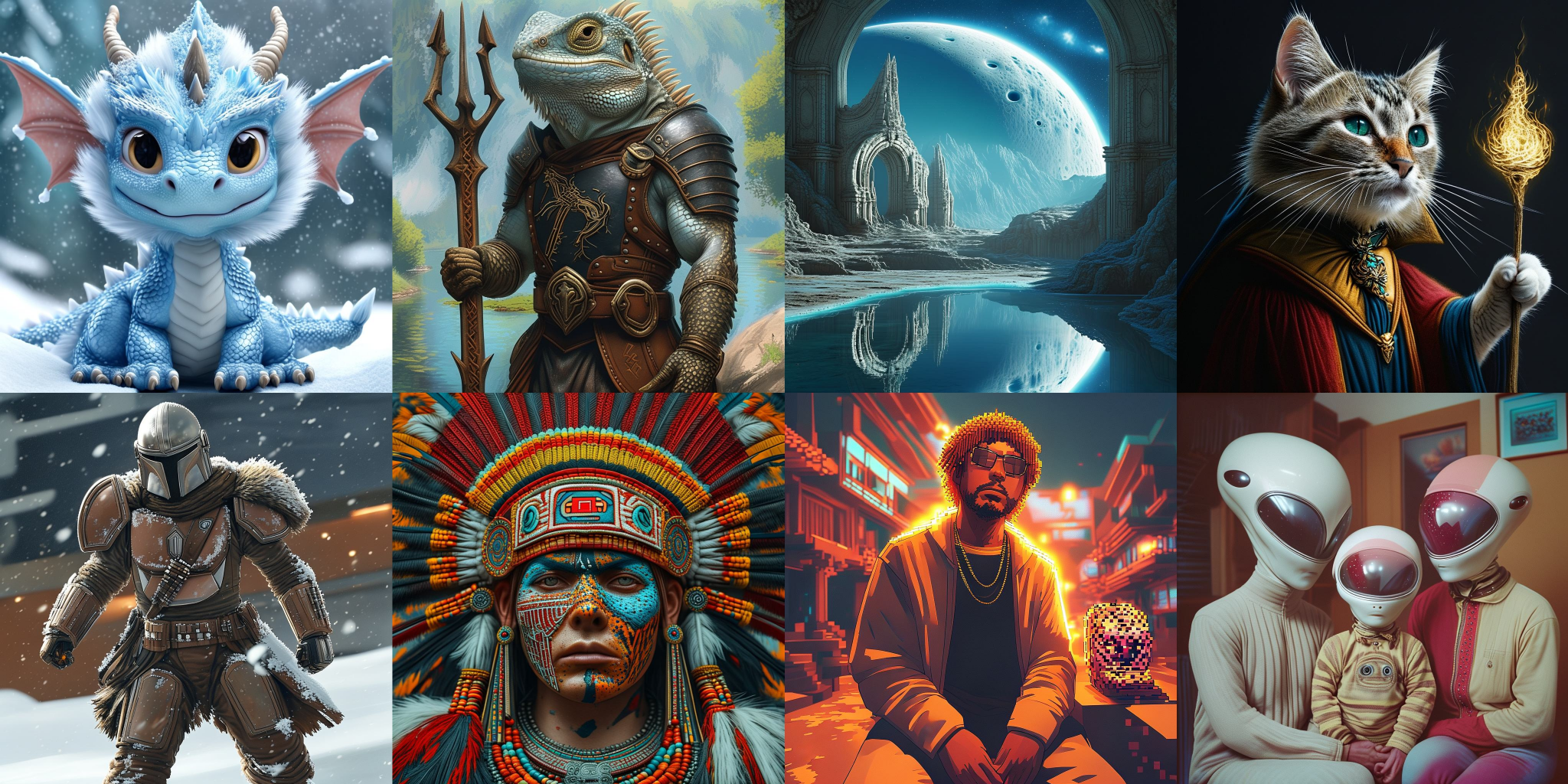}
    \vspace{-6mm}
    \captionof{figure}{\ourmethod produces high quality and aesthetic $1024{\times}1024$ image samples in around $0.5$ seconds.}
    \vspace{1mm}
    \label{fig:main}
\end{center}

}]

\begin{abstract}
\vspace{-2mm}
This work presents \ourmethod, a scale-wise transformer for text-to-image generation.
We start by adapting an existing next-scale prediction autoregressive (AR) architecture to T2I generation, investigating and mitigating training stability issues in the process.
Next, we argue that scale-wise transformers do not require causality and propose a non-causal counterpart facilitating ${\sim}21\%$ faster sampling and lower memory usage while also achieving slightly better generation quality.
Furthermore, we reveal that classifier-free guidance at high-resolution scales is often unnecessary and can even degrade performance.
By disabling guidance at these scales, we achieve an additional sampling acceleration of ${\sim}32\%$ and improve the generation of fine-grained details. 
Extensive human preference studies and automated evaluations show that \ourmethod outperforms existing T2I AR models and competes with state-of-the-art T2I diffusion models while being up to $7{\times}$ faster.
\end{abstract}

\section{Introduction}
\label{sec:intro}

Diffusion models (DMs)~\cite{song2019generative, song2020denoising,ho2020denoising,song2020score, karras2022elucidating, Karras2024edm2} are a dominating paradigm in visual content generation and have achieved remarkable performance in text conditional image~\cite{podell2024sdxl, flux, esser2024scaling, li2024playground}, video~\cite{polyak2024moviegencastmedia, videoworldsimulators2024} and 3D modeling~\cite{gao2024cat3d, melas20243d}.
Inspired by the unprecedented success of autoregressive (AR) models in natural language generation~\cite{dubey2024llama, touvron2023llamaopenefficientfoundation,geminiteam2024gemini15unlockingmultimodal}, numerous studies have focused on developing AR models specifically for visual content generation~\cite{li2024mar, fan2024fluid, tian2024var, lee2022autoregressiveimagegenerationusing, esser2020taming, sun2024autoregressive, Yu2022Pathways} to offer a more practical solution to the generative trilemma~\cite{xiao2022tackling}.

Traditional visual AR generative models perform \textit{next-token} prediction~\cite{esser2020taming, betker2023improving, sun2024autoregressive, lee2022autoregressiveimagegenerationusing, liu2024lumina-mgpt, wang2024emu3, liu2024lumina-mgpt, chameleonteam2024chameleon}.
These models flatten a 2D image into a 1D token sequence, and a causal transformer then predicts each token sequentially, resembling the text generation pipeline~\cite{Radford2018ImprovingLU, radford2019language, dubey2024llama, touvron2023llamaopenefficientfoundation}.
While this direction aims to unify vision and language modeling within a single AR framework, it still does not reach state-of-the-art diffusion models in terms of both speed and visual generation quality.

This discrepancy raises an important question: why do traditional AR models struggle in vision domains, whereas diffusion models excel?
\citet{tian2024var} and~\citet{chang2022maskgit} argue that next-token prediction imposes an unsuitable inductive bias for visual content modeling.
In contrast, diffusion models generate images in a coarse-to-fine manner~\cite{dieleman2024spectral, rissanen2023generative, baranchuk2022labelefficient}, a process that closely resembles human perception and drawing — starting with a global structure and gradually adding details.
Moreover, \citet{rissanen2023generative} and \citet{dieleman2024spectral} show that image diffusion models approximate spectral autoregression: progressively generating higher-frequency image components at each diffusion step.

Recently, scale-wise AR modeling has emerged as a natural and highly effective image generation solution via a \textit{next-scale} prediction paradigm~\cite{tian2024var, ma2024star, zhang2024varclip, tang2024hart}.
Unlike next-token prediction or masked image modeling~\cite{chang2022maskgit, li2024mar, fan2024fluid, li2023mage, chang2023muse}, scale-wise models start with a single pixel and progressively predict higher resolution versions of the image, while attending to the previously generated scales.
Therefore, next-scale prediction models perform coarse-to-fine image generation and may also share the spectral inductive bias observed in diffusion models~\cite{rissanen2023generative, dieleman2024spectral}, as upscaled images are generally produced by adding higher frequency details.
This makes scale-wise AR models a highly promising direction in visual generative modeling.
An important advantage of scale-wise models over DMs is that they perform most steps at lower resolutions, while diffusion models always operate at the highest resolution during the entire sampling process.
Therefore, scale-wise models yield significantly faster sampling while having the potential to provide similar generation performance to DMs~\cite{tian2024var}.

This work advances the family of scale-wise image generative models by introducing \ourmethod, a next-scale prediction transformer for text-to-image generation, that rivals the performance of diffusion models, while being significantly faster.
Drawing on recent developments~\cite{tian2024var, ma2024star}, we begin with implementing \ourmethodar, exploring and mitigating stability issues in the training process.

Then, we investigate whether the scale-wise AR models require attending to all previous scales. 
We notice that an input image at the current resolution already contains information about preceding scales by design.
Therefore, we hypothesize that the model may not need explicit attention to these levels within its architecture.
To test this, we remove the causal component from next-scale prediction transformers, enabling faster inference and improved scalability. 
In addition to the efficiency gains, we find that non-causal models deliver slightly better generative performance.

Also, we explore the influence of text conditioning across different resolution levels and observe that higher scales show minimal reliance on textual information. 
Leveraging this insight, we disable classifier-free guidance (CFG)~\cite{ho2022classifier} at the last scales, thereby reducing inference time by skipping extra forward passes required for CFG calculation. 
Interestingly, this not only accelerates sampling but also tends to mitigate generation artifacts in fine-grained details.

To sum up, the paper presents the following contributions: 

\begin{itemize}
    \item We introduce \ourmethod, a text-to-image next-scale prediction transformer that excludes explicit autoregression for more efficient sampling and better scalability. 
    As evidenced by human preference studies and automated evaluation, \ourmethod outperforms previous publicly available visual AR models.
    Compared to state-of-the-art text-to-image diffusion models, \ourmethod is up to $7{\times}$ faster while demonstrating competitive performance. 
    \item We demonstrate that using a non-causal transformer makes \ourmethod ${\sim}21\%$ more efficient for $1024{\times}1024$ image generation due to cheaper attention operations. 
    Additionally, \ourmethod reduces memory consumption during inference, previously needed for storing key-value (KV) cache, by up to $2.3$ GB for a single image, enabling better scaling to higher resolution image generation.
    Moreover, \ourmethod slightly surpasses its AR counterpart in generation quality under the same training setups.
    \item We find that \ourmethod has weaker reliance on the text at high resolution scales. 
    This observation allows us to disable classifier-free guidance on the last two steps, resulting in further ${\sim}32\%$ acceleration and better generation of fine-grained details, as confirmed by human evaluation.
\end{itemize}

\section{Related work}
\label{sec:related_work}

\subsection{Text-to-image diffusion models}

Text-conditional diffusion models (DMs)~\cite{flux, recraft, esser2024scaling, saharia2022photorealistic, podell2024sdxl, gao2024lumina-next, betker2023improving} have become the de facto solution for text-to-image (T2I) generation. 
Despite their impressive performance, a well-known limitation of DMs is slow sequential inference, which hampers real-time or large-scale generation tasks.
Most publicly available state-of-the-art T2I diffusion models~\cite{flux, podell2024sdxl, esser2024scaling, rombach2021highresolution, gao2024lumina-next, li2024playground} operate in the VAE~\cite{vae2014} latent space, allowing for more efficient sampling of high-resolution images. 
However, these models still require $20{-}50$ diffusion steps in the latent space.

Diffusion distillation methods~\cite{meng2023distillation, song2023consistency, luo2023latent, sauer2023adversarial, sauer2024fast, yin2024onestep, yin2024improved, kim2024pagodaprogressivegrowingonestep} are the most promising direction for reducing the number of diffusion steps to just $2{-}4$. 
Current state-of-the-art approaches, such as DMD2~\cite{yin2024improved} and ADD~\cite{sauer2023adversarial}, demonstrate strong generation performance in $4$ steps and may even surpass the teacher performance in terms of image quality thanks to additional adversarial training on real images. 

\subsection{Visual autoregressive modeling}

Autoregressive (AR) models is a promising alternative paradigm for image generation that can be categorized into three main groups: next-token prediction~\cite{sun2024autoregressive, esser2020taming, Yu2022Pathways, lee2022autoregressiveimagegenerationusing}, next-scale prediction~\cite{tian2024var, ma2024star, tang2024hart}, and masked autoregressive models~\cite{li2024mar, fan2024fluid}.

Next-token prediction AR models are similar to GPT-like causal transformers~\cite{Radford2018ImprovingLU, radford2019language, dubey2024llama, touvron2023llamaopenefficientfoundation} and generate an image token by token using some scanning strategy, e.g., raster order (left to right, top to bottom).
The tokens are typically obtained using VQ-VAE-based discrete image tokenizers~\cite{vqvae, esser2020taming, yu2022vectorquantized, lee2022autoregressiveimagegenerationusing}. 
VQ-VAE maps an image to a low-resolution 2D latent space and assigns each latent "pixel" to an element in the learned vocabulary.

Masked autoregressive image modeling (MAR)~\cite{li2024mar, fan2024fluid} extends masked image generative models~\cite{chang2022maskgit, li2023mage, chang2023muse} and predicts multiple masked tokens in random order at a single step. 
Notably, MAR operates with continuous tokens, using a diffusion loss for training and a lightweight token-wise diffusion model for token sampling. 
Fluid~\cite{fan2024fluid} applies this approach to T2I generation and explores its scaling behavior.

Next-scale prediction AR modeling, introduced by VAR~\cite{tian2024var}, represents an image as a sequence of scales of different resolutions. 
Unlike next-token prediction and masked AR modeling, the scale-wise transformer predicts all tokens at a higher resolution in parallel, attending to previously generated lower-resolution scales.

To represent an image with a sequence of scales, VAR~\cite{tian2024var} uses a hierarchical VQ-VAE that maps an image to a pyramid of latent variables of different resolutions (scales), progressively constructed using residual quantization (RQ)~\cite{lee2022autoregressiveimagegenerationusing}. 
In the following, we will refer to this VAE model as \textit{RQ-VAE}. 
Each latent variable in RQ-VAE is associated with a set of discrete tokens from a shared vocabulary across all scales, similar to a single-layer VQ-VAE.

During sampling, scale-wise AR model $\theta$ iteratively predicts image tokens scale-by-scale, formulated as:
\[p_\theta(s_1, \dots, s_N|c) = \prod_{i=1}^{N} p_\theta(s_i|s_1, \dots, s_{i-1}, c),\]
where $s_i$ represents RQ-VAE tokens at the current scale, $N$ is the total number of scales, and $c$ is the conditioning information. 
The model is a transformer~\cite{radford2019language} with a block-wise causal attention mask, as shown in \Cref{fig:attention_mask} (Left). 
VAR~\cite{tian2024var} adopts a transformer architecture from DiT~\cite{Peebles2022DiT}.

Recent works have applied next-scale prediction models to T2I generation~\cite{ma2024star, zhang2024varclip, tang2024hart}. 
STAR~\cite{ma2024star} uses the pretrained RQ-VAE model from VAR and modifies its generator to effectively handle text conditioning. 
Although STAR has not been released as of the writing of this paper, we consider STAR as our baseline architecture, from which we gradually progress towards the proposed model, \ourmethod.

A concurrent work, HART~\cite{tang2024hart}, proposes a lightweight T2I scale-wise AR model with only $0.7$B parameters.
It mainly addresses the limitations of the discrete RQ-VAE in VAR by introducing an additional diffusion model to model continuous error residuals, resulting in a hybrid model: a scale-wise AR model combined with a diffusion model for refining the reconstructed latents.
In contrast, we focus solely on designing the scale-wise generative transformer using the pretrained RQ-VAE from VAR~\cite{tian2024var}, slightly tuning it for $512{\times}512$ resolution.
Combining our scale-wise generative model design with HART's hybrid tokenization could be a promising direction for future work.

MAR~\cite{li2024mar} can also be considered as a hybrid model that combines both autoregressive and diffusion priors. 
Disco-diff~\cite{xu2024discodiff} conditions a diffusion model on discrete tokens produced with a next-token prediction transformer.
DART~\cite{gu2024dart} introduces AR transformers as a backbone for non-markovian diffusion models.
Opposed to this line of works, \ourmethod does not use any diffusion prior.

\section{Method}
\label{sec:method}

\subsection{Architecture}
\label{sec:basic_architecture}

As a starting point, we design a basic text-conditional architecture closely following VAR~\cite{tian2024var} and STAR~\cite{ma2024star}.
Scale-wise AR text-to-image generation pipeline comprises three main components: RQ-VAE~\cite{lee2022autoregressiveimagegenerationusing} as an image tokenizer, a pretrained text encoder~\cite{Radford2021LearningTV}, and a scale-wise block-wise causal transformer~\cite{tian2024var}.

Our model adopts the pretrained RQ-VAE from VAR~\cite{tian2024var}, which represents an image with $N{=}10$ scales for resolutions $256$ and $512$, and $N{=}14$ for $1024$.
We slightly tune RQ-VAE on $512{\times}512$ resolution, as discussed in \Cref{sec:vae_tuning}.

To ensure strong image-text alignment of the resulting model, we follow the literature in T2I diffusion modeling~\cite{podell2024sdxl,esser2024scaling} and employ two text encoders: CLIP ViT-L~\cite{Radford2021LearningTV} and OpenCLIP ViT-bigG~\cite{ilharco_gabriel_2021_5143773}.
The text embeddings extracted from each model are concatenated along the channel axis.

The transformer block architecture is adopted from VAR~\cite{tian2024var}, where we incorporate cross-attention~\cite{vaswani2017attention} layers between a self-attention layer and feed-forward network (FFN) in each transformer block.
A pooled embedding from OpenCLIP ViT-bigG is propagated to the transformer blocks via Adaptive Layer Normalization (AdaLN)~\cite{xu2019understanding}.

We also incorporate conditioning on the cropping parameters following SDXL~\cite{podell2024sdxl} to mitigate unintended object cropping in generated images.
Specifically, we transform center crop coordinates $c_{top}, c_{left}$ into Fourier feature embeddings and concatenate them.
Then, we map the obtained vector to the hidden size of OpenCLIP ViT-bigG via a linear layer and add it to the pooled embedding.

RMSNorm~\cite{zhang2019root} layers are applied to the inputs of attention and FFN blocks.
Layer-normalization (LN) layers~\cite{ba2016layernormalization} are used for query-key (QK) normalization. 
We also use normalized 2D rotary positional embeddings (RoPE)~\cite{ma2024star, heo2024ropevit}, which allow faster model adaptation to higher resolutions.
We use SwiGLU activation function~\cite{shazeer2020glu} in FFN blocks.

\begin{figure}
\centering
\includegraphics[width=0.93\linewidth]{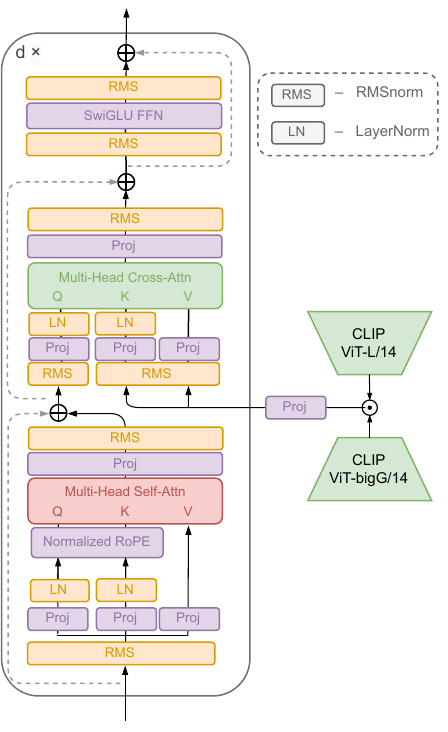}
\vspace{-4mm}
\caption{Transformer block in the \ourmethod model.}
\label{fig:architecture}
\vspace{-1em}
\end{figure}

\subsection{Training stability issues}
\label{sec:our_architecture}

Here, we analyze the training performance of the basic model with $d{=}20$ transformer blocks and introduce a simple modification to improve training stability.

We train the model in mixed-precision BF16/FP32 for $150$K iterations on the $256{\times}256$ image-text dataset described in \Cref{sec:pretraining}.
The detailed training and evaluation setups are presented in \Cref{app:training_details_analysis}.
During the training, we track activation norms and standard metrics, such as FID~\cite{heusel2017gans}, CLIP Score~\cite{clip_score} and PickScore~\cite{Kirstain2023PickaPicAO}.

\begin{figure}[t]
\centering
\includegraphics[width=\linewidth]{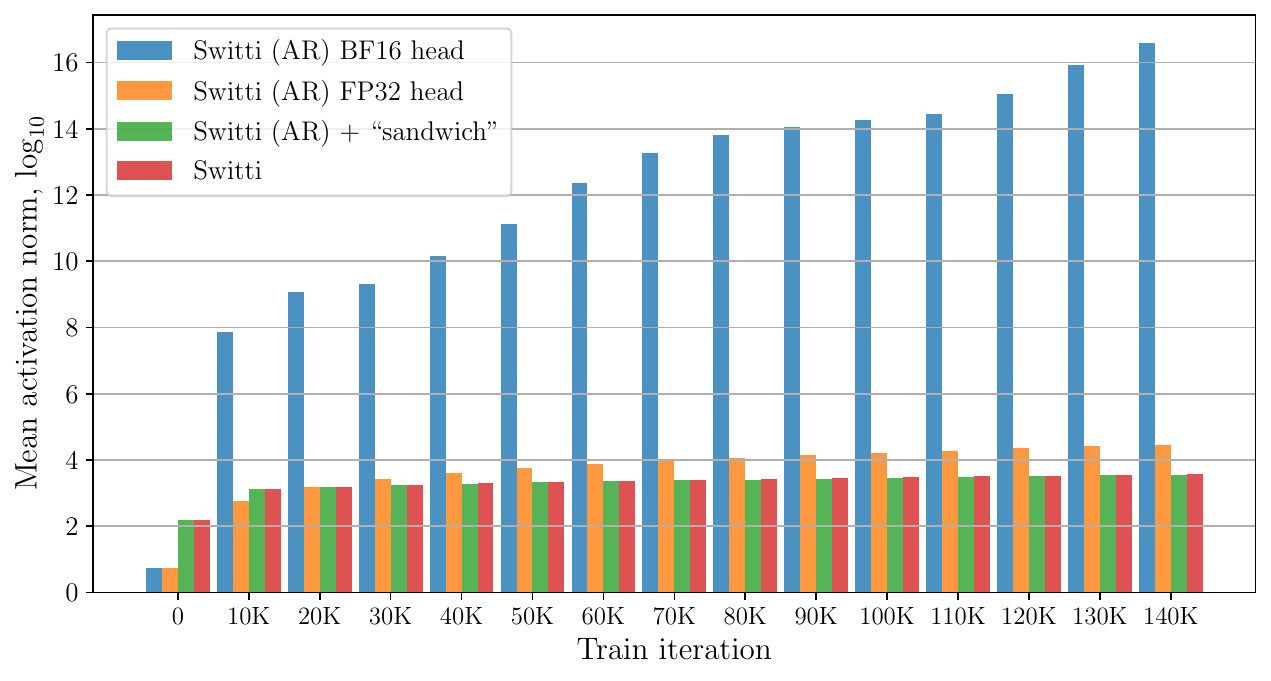}
\vspace{-20pt}
\caption{
    Last transformer block activation norms over training.
    Casting the prediction head to full-precision reduces the norm growth. ``Sandwich''-normalization further mitigates the issue.
}
\label{fig:activations_growth}
\end{figure}

First, we observe stability issues during training, leading to eventual divergence in our large scale experiments or suboptimal performance.
Our investigation reveals that the root of this issue lies in the rapid growth of transformer activation norms, as illustrated in~\Cref{fig:activations_growth} (Blue). 
Activation norms of the last transformer block grow throughout training iterations, reaching extremely large values of $10^{16}$.

\begin{figure}[t]
    \centering
    \includegraphics[width=\linewidth]{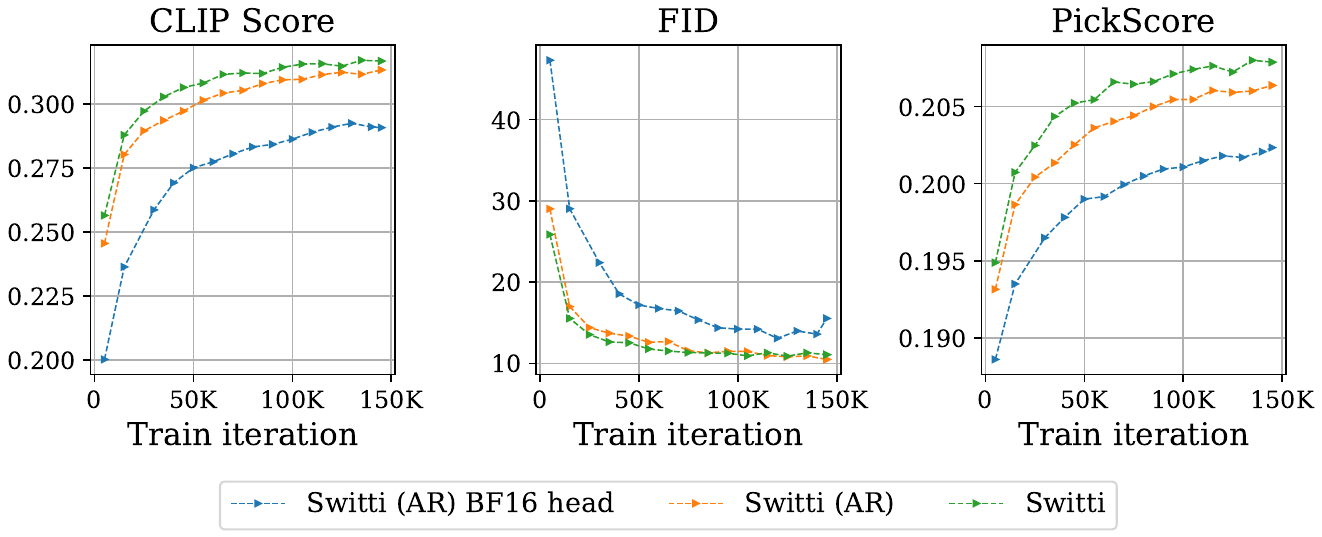}
    \vspace{-15pt}
    \caption{Evaluation of $d{=}20$ models on COCO 30K.
    Using the non-causal attention mask also slightly improves the performance.
    }
    \label{fig:activations_architectures}
    \vspace{-2mm}
\end{figure}

Therefore, the first step towards stabilizing the training is to cast the model head to FP32 during training.
We find this as a critical technical detail that significantly reduces activation norms, resulting in much better convergence, as we show in~\Cref{fig:activations_architectures} (Orange).
However, this trick does not fully address the problem since activation norms still keep growing and reach high values of $10^{4}$ by the end of training.

To further reduce the growth of activation norms during training, we employ ``sandwich''-like normalizations~\cite{ding2021cogview, gao2024lumina-next}, to keep the activation norms in a reasonable range.
Specifically, we insert additional normalization layers right after each attention and feed-forward blocks.
As we show in~\Cref{fig:activations_growth}, this modification further mitigates the growth of activation norms during training.
However, we find that neither ``sandwdich'' normalizations nor the choice of normalization functions do not affect the metrics of the models at this scale, as evidenced in~\Cref{app:normalizations}.

We illustrate the transformer block of the described architecture in~\Cref{fig:architecture} and denote the scale-wise AR model with the proposed architecture as \ourmethodar.

\begin{figure}[h]
    \centering
    \includegraphics[width=\linewidth]{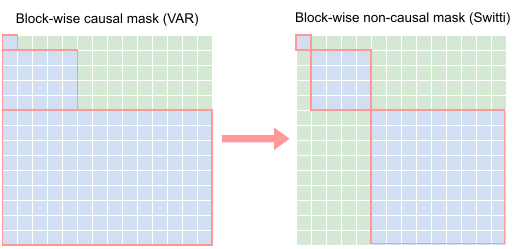}
    \caption{Visualization of the block-wise self-attention masks in VAR (Left) and \ourmethod (Right).}
    \label{fig:attention_mask}
\end{figure}

\subsection{Employing a non-causal transformer}
\label{sec:self_attention_maps}
Next, we delve into the autoregressive next-scale prediction sampling of the original VAR~\cite{tian2024var}.
At each generation step $i$, the VAR transformer predicts a sequence of tokens $s_i$ conditioned on the previously generated tokens $s_0, \dots, s_{i-1}$. %
The embeddings of these tokens form a feature map of the spatial size ${h_i{\times}w_i}$.
This feature map is then upscaled to $h_{i+1}{\times}w_{i+1}$ and combined with the previously predicted feature maps to serve as an input for the next step.

\begin{figure}[t]
    \centering
    \includegraphics[width=\linewidth]{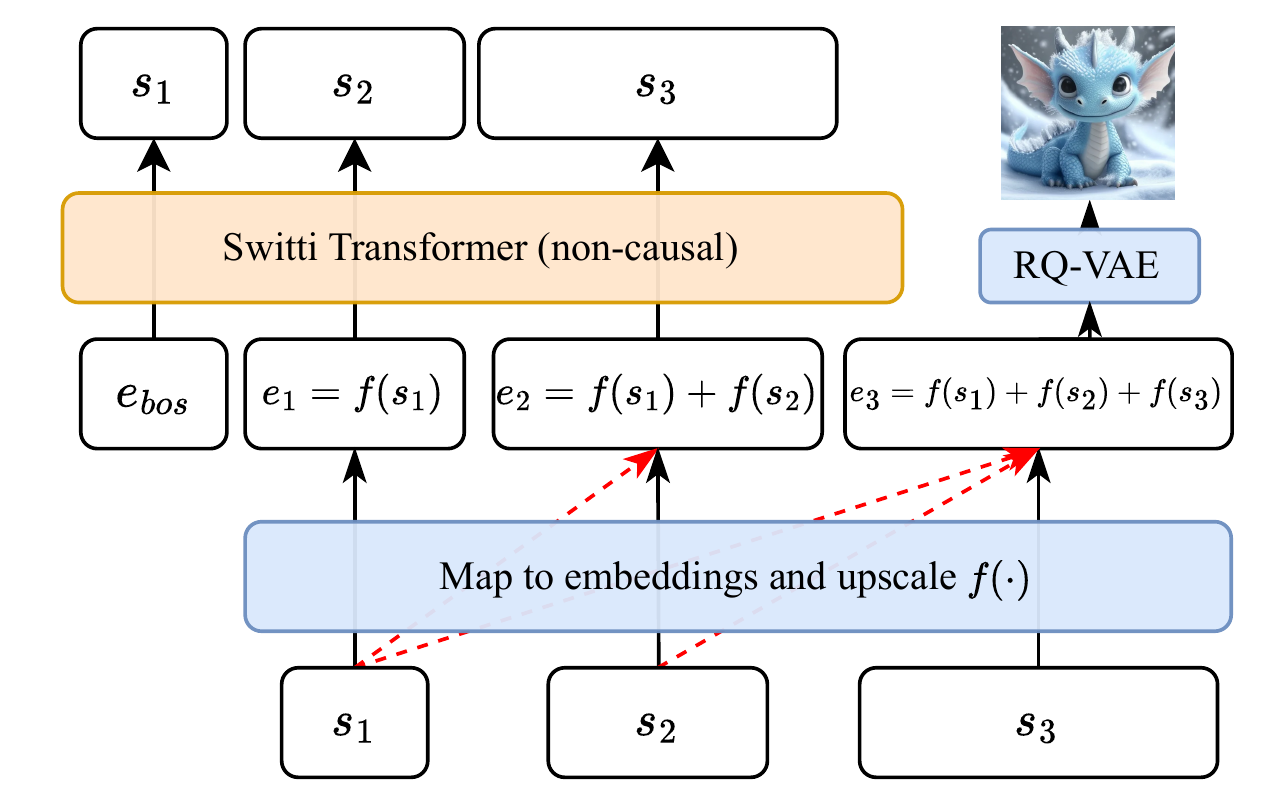}
    \vspace{-12pt}
    \caption{
        Sampling with \ourmethod. 
        Model inputs at each scale already incorporate representations of previously generated scales, motivating us to employ a non-causal transformer.
    }
    \label{fig:non_causal_inference}
    \vspace{-15pt}
\end{figure}

Therefore, we observe that the conditioning on the preceding scales occurs twice in VAR's image generation: first, implicitly, when forming the model input, and second, explicitly, using the causal transformer.
Based on this observation, we update the attention mask to allow self-attention layers to attend only to tokens at the current scale, as shown in~\Cref{fig:attention_mask} (Right). 
This implies that the transformer is no longer causal, enabling more efficient sampling due to cheaper attention operations and eliminating the need for key-value (KV) cache.
\Cref{fig:non_causal_inference} illustrates the next-scale prediction sampling using a non-causal transformer.
Overall, we refer to our scale-wise model with the non-causal transformer architecture as \ourmethod.

A concurrent work, HART~\cite{hart}, makes a similar observation that self-attention in VAR is mostly local, allowing them to discard up to 80\% tokens during training, while the model remains auto-regressive.
In contrast, we remove causality in transformer and observe that this modification slightly improves performance in terms of CLIP Score and PickScore, as we show in~\Cref{fig:activations_architectures} (Green).
\Cref{app:non_causal_loss} provides additional training loss analysis to justify our choice.

\subsection{The role of text conditioning}
\label{sec:cross_attention_maps}

Finally, we examine the effect of text conditioning at different model scales. 

\begin{figure}[tb]
    \centering
    \includegraphics[width=\linewidth]{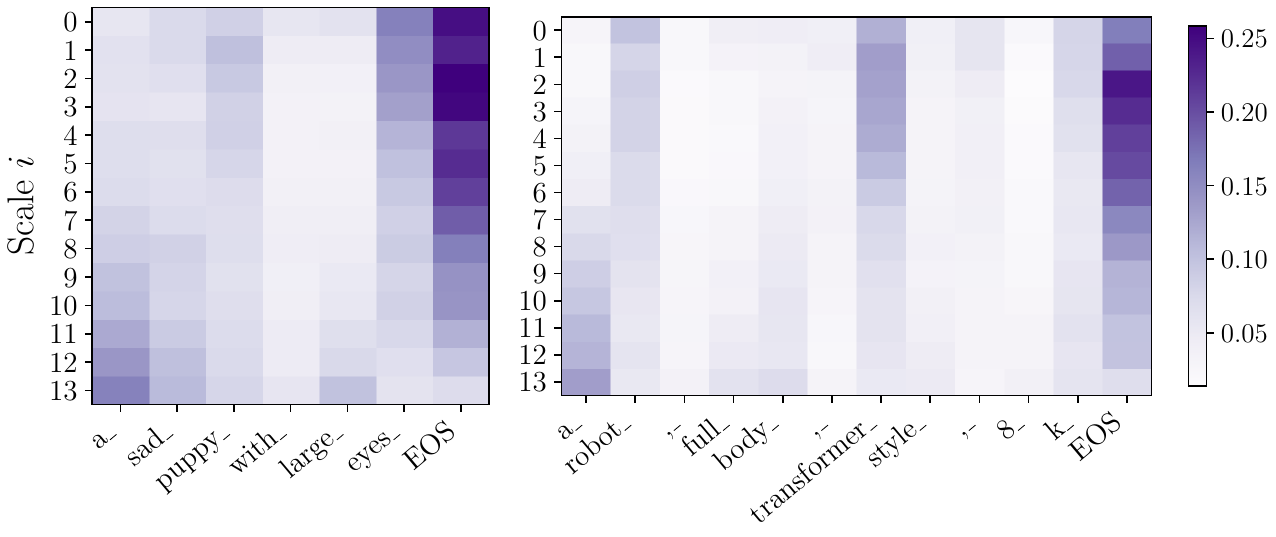}
    \vspace{-15pt}
    \caption{
        Cross-attention maps between image tokens at each scale (Y-axis) and text tokens (X-axis).
        Attention scores decrease for main semantic tokens and pooled embeddings (EOS), while increase for less informative tokens, e.g., articles.
    }
    \label{fig:cross_attention}
    \vspace{-0.5em}
\end{figure}

\paragraph{Cross-attention.}
First, we plot a cross-attention map between image tokens at different scales and text tokens (except the BOS-token), averaged across transformer blocks.
We find that attention scores to the BOS-token are high for all scales and omit it to better visualize the pattern for other tokens.
Figure~\ref{fig:cross_attention} shows a typical pattern for two random prompts. 
At higher scales, the scores decrease for the EOS-token (used as a pooled embedding) but increase for the less semantic tokens, such as articles and stylistic words.
This suggests a weaker reliance on the prompt at later scales.
Such behavior is consistent across different prompts.

\paragraph{Prompt switching.}
Next, we investigate the impact of text-conditioning at various scales by switching the text prompt to a new one starting from a certain scale. 
The visualization of prompt switching is presented in Figure~\ref{fig:switch_prompt_main}, with additional examples provided in Appendix~\ref{app:switch_prompt}.
Indeed, the prompt has minimal influence on the image semantics at the last two scales.
Interestingly, switching a prompt at the middle scales results in a simple image blending approach.

\paragraph{Practical implication.}
Classifier-free guidance (CFG)~\cite{ho2022classifier} is an important sampling technique for high-quality text-conditional generation that requires an extra model forward pass at each step. 
Specifically for the scale-wise models, calculations at the last scales take up most of the computing time of the entire sampling process. 
To save costly model forward passes in high resolution, we propose disabling CFG at the last scales, expecting little effect on generation performance,
as also was recently noted in diffusion models~\cite{kynkaanniemi2024applying}.

\begin{figure}[t]
    \centering
    \includegraphics[width=1.0\linewidth]{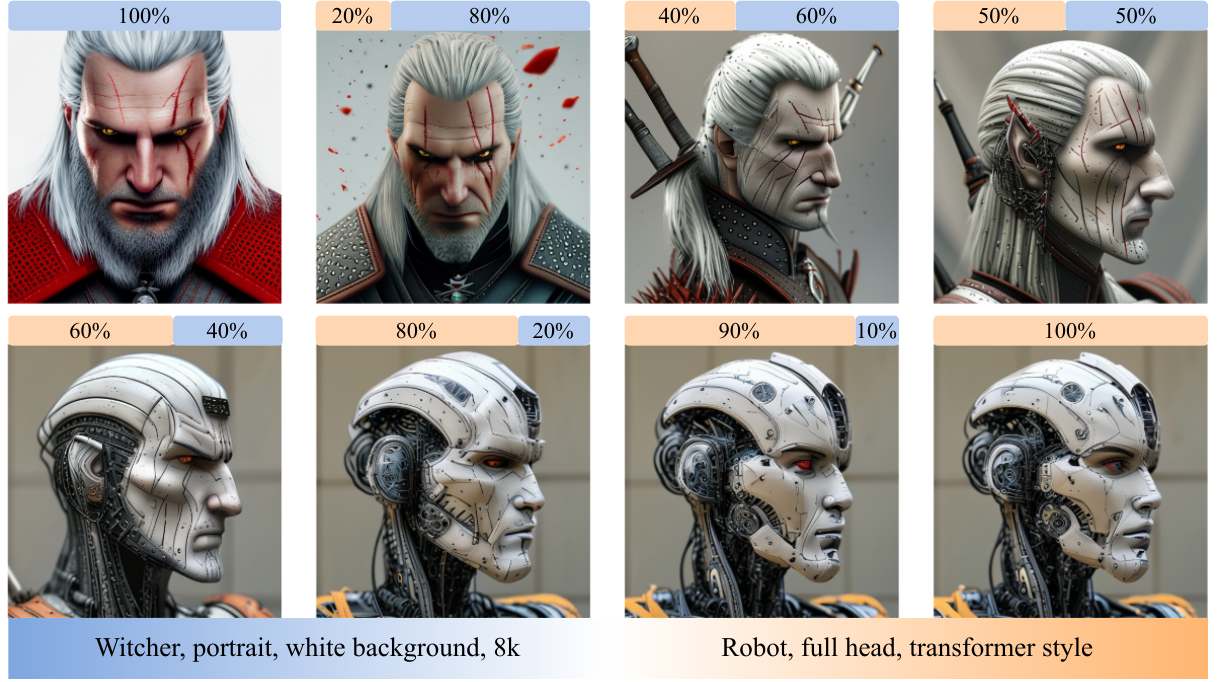}
    \caption{
        Prompt switching within $512{\times}512$ generation.
        Changing the prompt at late steps has a minor effect on the resulting image.
    }
    \label{fig:switch_prompt_main}
    \vspace{-1em}
\end{figure}

\section{Model Training}
\label{sec:training}

\subsection{Pretraining}
\label{sec:pretraining}

Following the transformer scaling setup in VAR~\cite{tian2024var}, we set the number of transformer layers $d{=}30$ for our main model, resulting in ${\sim}2.5$B trainable parameters.
We pretrain the model in two stages: first on images of $256{\times}256$ resolution, followed by training at $512{\times}512$.

As a training dataset, we collect $100$M image-text pairs that are filtered from a base set of $6$B pairs from the web. 
Images are filtered to ensure high aesthetic quality and recaptioned using open-source VLMs.
We provide more technical details about data filtering and training hyperparameters in~\Cref{app:training_details_analysis}.

\subsection{Supervised fine-tuning}
\label{sec:sft}
After the pretraining phase, we further fine-tune the model using ${\sim}40,000$ text-image pairs, inspired by practices in large-scale T2I diffusion models~\cite{kastryulin2024yaart,liu2024playgroundv3, li2024playground}.
The pairs are manually selected by assessors instructed to capture exceptionally aesthetic, high quality images with highly relevant and detailed textual descriptions.

The model is fine-tuned for $40$K iterations with a batch size of $64$ and a learning rate $5e{-}7$ on image central crops of resolution $1024{\times}1024$.

In addition, we slightly perturb the RQ-VAE latents prior to the quantization step with Gaussian noise ($\sigma{=}0.01$) as an augmentation to mitigate overfitting.
We observe that the perturbed latents after the quantization and dequantization steps produce visually indistinguishable images while resulting in ${\sim}80\%$ different tokens in a sequence, indicating that tokens at latter scales have less impact on the image quality.

\subsection{RQ-VAE tuning}
\label{sec:vae_tuning}

In this work, we use the released RQ-VAE from VAR~\cite{tian2024var} that was trained on $256{\times}256$ images and fine-tune its decoder to adapt it for $512{\times}512$ resolution. 
The encoder and codebook are kept frozen to preserve the same latent space, allowing to fine-tune the autoencoder independently from the generator. 
Following the standard practice in super-resolution literature~\cite{ledig2017srgan}, we use a combination of $L_{1}$ reconstruction, LPIPS perceptual~\cite{zhang2018perceptual} and adversarial~\cite{ledig2017srgan} losses, resulting in the following fine-tuning objective:
\begin{equation}
\mathcal{L} = \mathcal{L}_{L_1} + \mathcal{L}_{\mathrm{LPIPS}} + \mathcal{L}_{\mathrm{adv}}
\end{equation}
We adopt a UNetSN discriminator from~\citet{wang2021realesrgan} for adversarial loss.
The generator and discriminator are trained for $100$K steps with a batch size of $256$ and a constant learning rate of $1e{-}5$.

To compare the reconstruction quality of the original RQ-VAE and the one with a tuned decoder, we compute classic full-reference metrics PSNR, SSIM~\cite{ssim}, LPIPS~\cite{zhang2018perceptual} and no-reference CLIP-IQA~\citep{wang2022exploring} metric on a held-out dataset of 300 images.
Results in~\Cref{tab:metrics_vae} demonstrate that the fine-tuned RQ-VAE outperforms an original model with respect to all metrics.
Interestingly, the gains from fine-tuning RQ-VAE decoder for $512{\times}512$ resolution automatically transferred to $1024{\times}1024$ image reconstruction.
Additionally, we provide several representative comparisons in~\Cref{app:vae_visual_comparison}.

\begin{table}[htb]
\small
\centering
\begin{tabular}{lcccc}
\toprule
Model & PSNR $\uparrow$ & SSIM $\uparrow$ & LPIPS $\downarrow$ & CLIP-IQA $\uparrow$ \\
\midrule
\multicolumn{5}{c}{$512{\times}512$ Reconstruction} \\
\midrule
Original & 21.60 & 0.634  & 0.200	& 0.727 \\
Fine-tuned & 22.27 & 0.653  & 0.188	& 0.772 \\
\midrule
\multicolumn{5}{c}{$1024{\times}1024$ Reconstruction} \\
\midrule
Original & 22.53 & 0.703 & 0.207  & 0.683 \\
Fine-tuned & 23.91 & 0.733  & 0.177 & 0.748 \\
\bottomrule
\end{tabular}
\vspace{-3mm}
\caption{Comparison of a fine-tuned and an original RQ-VAE.}
\vspace{-5pt}
\label{tab:metrics_vae}
\end{table}

We believe that more pronounced gains can be achieved via more thorough investigation of RQ-VAE and training the entire model. 
We leave this direction for future work.

\section{Experiments}

\begin{table*}[ht]
\small
\setlength\tabcolsep{3pt}
\renewcommand{\arraystretch}{0.9}
\centering
\begin{tabular}{lcc|cccc|cccc|c}
\toprule
 & & & \multicolumn{4}{c}{COCO 30K eval prompts} & \multicolumn{4}{c}{MJHQ 30K eval prompts} & \\
\midrule
Model & \begin{tabular}{@{}c@{}}Latency,\\s/image \end{tabular} & \begin{tabular}{@{}c@{}}Parameters\\count, B\end{tabular} & PickScore $\uparrow$ & CLIP $\uparrow$ & IR $\uparrow$ & FID $\downarrow$ & PickScore $\uparrow$ & CLIP $\uparrow$ & IR $\uparrow$ & FID $\downarrow$ & GenEval $\uparrow$ \\
\midrule
\multicolumn{12}{c}{Distilled Diffusion Models} \\
\midrule
SDXL-Turbo~\cite{sauer2023adversarial} & 0.4 & 2.6 & \textcolor{blue}{0.229} & \textcolor{olive}{0.355} & 0.83 & \textcolor{blue}{17.6} & \textcolor{olive}{0.216} & 0.365 & \textcolor{olive}{0.84} & 15.7 & 0.55 \\
DMD2~\cite{yin2024improved} & 0.4 & 2.6 & \textcolor{red}{0.231} & \textcolor{blue}{0.356} & \textcolor{olive}{0.87} & \textcolor{red}{14.3} & \textcolor{red}{0.219} & 0.374 & \textcolor{blue}{0.87} & \textcolor{blue}{7.2} & \textcolor{olive}{0.58} \\
\midrule
\multicolumn{12}{c}{Diffusion Models} \\
\midrule
SDXL~\cite{podell2024sdxl} & 2.3 & 2.6 & 0.226 & \textcolor{red}{0.360} & 0.77 & \textcolor{red}{14.4} & \textcolor{blue}{0.217} & \textcolor{red}{0.384} & 0.78 & \textcolor{olive}{7.6} & 0.55 \\
SD3-medium~\cite{esser2024scaling} & 3.9 & 2.0 & \textcolor{olive}{0.227} & 0.354 & \textcolor{red}{1.01} & 19.5 & 0.215 & 0.363 & \textcolor{red}{0.91} & 13.1 & \textcolor{red}{0.65} \\
Lumina-Next~\cite{gao2024lumina-next} & 5.8 & 2.0 & 0.224 & 0.329 & 0.55 & 18.4 & \textcolor{olive}{0.216} & 0.353 & 0.75 & \textcolor{red}{5.9} & 0.47 \\
\midrule
\multicolumn{12}{c}{Autoregressive Models} \\
\midrule
LlamaGen~\cite{sun2024autoregressive} & 3.8 & 0.8 & 0.208 & 0.274 & -0.25 & 44.8 & 0.194 & 0.288 & -0.45 & 26.9 & 0.32 \\
HART~\cite{tang2024hart} & 0.5 & 0.7 & 0.223 & 0.341 & 0.75 & 20.9 & \textcolor{olive}{0.216} & 0.366 & \textcolor{olive}{0.84} & \textcolor{red}{5.8} & 0.55 \\
\midrule
\ourmethod 512 (ours) & 0.1 & 2.5 & \textcolor{olive}{0.227} &	\textcolor{blue}{0.356} & \textcolor{blue}{0.95}	& \textcolor{blue}{17.6}	&\textcolor{blue}{0.217} & \textcolor{blue}{0.381} & \textcolor{red}{0.91} & 9.5 & \textcolor{blue}{0.62} \\
\ourmethod 1024 (ours) & 0.5 & 2.5 & \textcolor{blue}{0.229} &	\textcolor{olive}{0.355} & \textcolor{blue}{0.96} & \textcolor{olive}{18.2}	&\textcolor{red}{0.219} & \textcolor{olive}{0.378} & 0.81 & 8.1 & \textcolor{blue}{0.62} \\
\bottomrule
\end{tabular}

\vspace{-3mm}
\caption{Quantitative comparison of \ourmethod to other competing open-source models. The best model is highlighted in \textcolor{red}{red}, the second-best in \textcolor{blue}{blue}, and the third-best in \textcolor{olive}{yellow} according to the respective automated metric.}
\label{tab:metrics_baselines}
\vspace{-10pt}
\end{table*}

\begin{figure*}[ht]
    \centering
    \includegraphics[width=\linewidth]{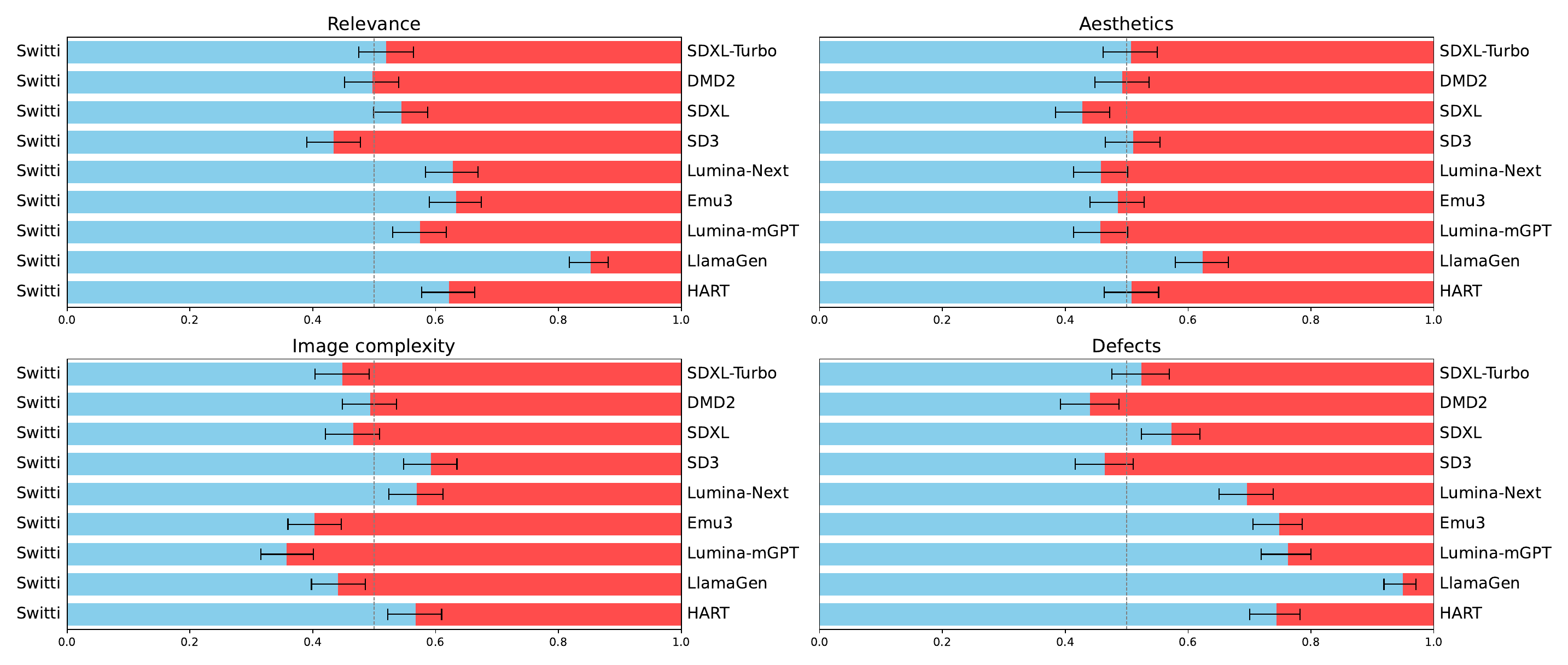}
    \vspace{-20pt}
    \caption{Human study comparing \ourmethod with competing AR, diffusion-based models. Error bars correspond to a 95\% confidence interval.}
    \vspace{-9pt}
    \label{fig:sbs_baselines}
\end{figure*}

\begin{figure*}[ht]
\centering
\includegraphics[width=\linewidth]{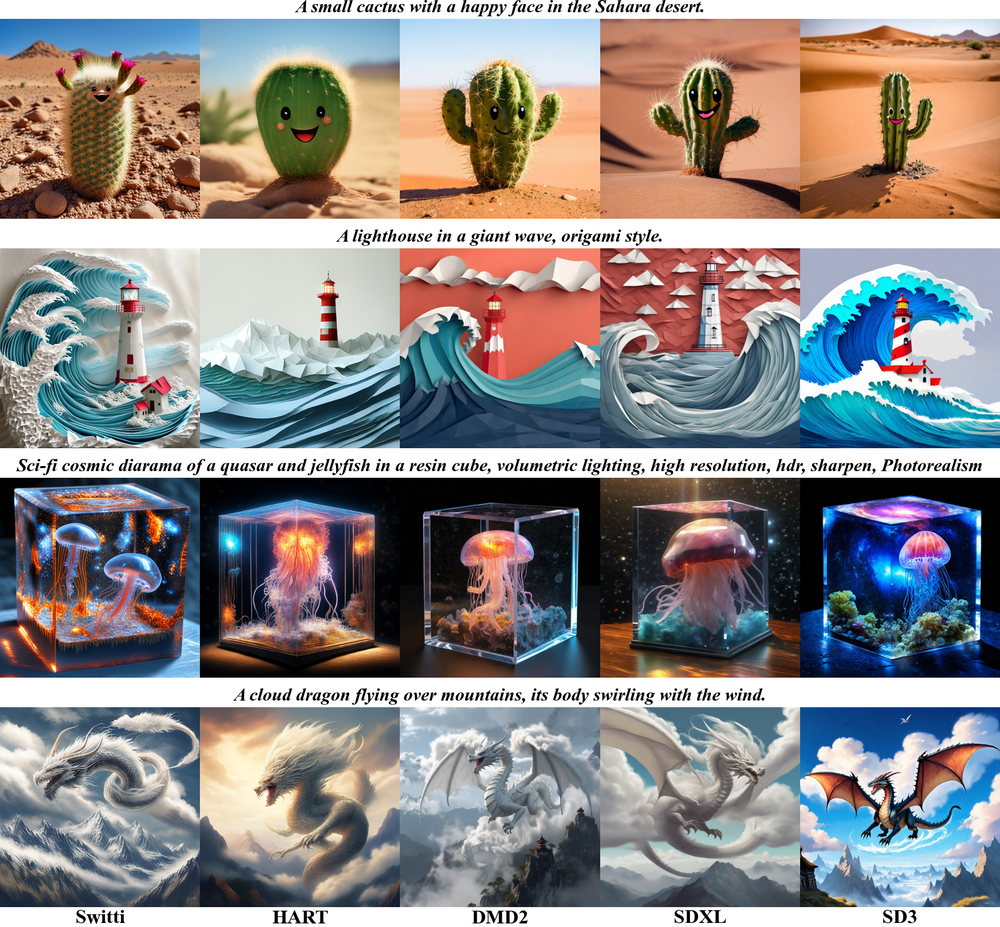}
\caption{
    Qualitative comparison of \ourmethod for $1024{\times}1024$ generation against the baselines. 
    More examples are in~\Cref{app:visual_comparison}.}
\label{fig:model_comparison_2}
\end{figure*}

We compare our final models with several competitive text-to-image baselines from various generative families:
\begin{itemize}
    \item \textbf{Diffusion models}: Stable Diffusion XL~\cite{podell2024sdxl}, Stable Diffusion 3 Medium~\cite{esser2024scaling}, Lumina-Next~\cite{gao2024lumina-next}.
    \item \textbf{Diffusion distillation}: SDXL-Turbo~\cite{sauer2023adversarial}, DMD2~\cite{yin2024improved}
    \item \textbf{Autoregressive models}: Emu3~\cite{wang2024emu3}, Lumina-mGPT~\cite{liu2024lumina-mgpt}, 
    LlamaGen-XL~\cite{sun2024autoregressive}, HART~\cite{tang2024hart}.
\end{itemize}

\subsection{Automated evaluation}

\paragraph{Evaluation setting.}
To comprehensively evaluate the performance of our models, we use a combination of established automated metrics and a human preference study.
For automated metrics, we report CLIP Score~\cite{clip_score}, ImageReward~\cite{xu2023imagereward}, PickScore~\cite{Kirstain2023PickaPicAO}, FID~\cite{heusel2017gans} and GenEval~\cite{geneval}.
For all evaluated models, we generate images in their native resolution and resize them to $512{\times}512$.
More details about the evaluation pipeline are provided in~\Cref{app:evaluation details}.

We calculate metrics on two validation datasets frequently used for text-to-image evaluation: MS-COCO~\cite{ms_coco} and MJHQ~\cite{li2024playground}.
For both datasets, we generate one image for each of $30,000$ validation prompts.

\paragraph{Results.}
We present the metric values in \Cref{tab:metrics_baselines}. 
\ourmethod achieves comparable performance to the baselines, ranking top-3 for 7 out of 9 automated metrics while exhibiting higher efficiency than most competitors. 
We do not provide automated metrics for Emu3 and Lumina-mGPT due to their exceptionally long sampling times, that makes generating 60,000 images infeasible in a reasonable time.

\subsection{Human evaluation}
While automated metrics are widely adopted for evaluating text-to-image models, we argue that they do not fully reflect all aspects of image generation quality.
Therefore, we conduct a rigorous human preference study across several aspects: the presence of defects, textual alignment, image complexity and aesthetics quality.
\Cref{app:human_eval} provides more details about the human evaluation setup.
For human evaluation, we generate four images for each of $128$ captions from Parti prompts collection~\cite{parti}, a set specifically curated for human preference study~\cite{esser2024scaling, yin2024improved, sauer2023adversarial}.

\paragraph{Sampling setup.}
For a side-by-side comparison, we generate images with classifier-free guidance set to 6 and deactivate it at the last two scales, as described in~\Cref{sec:cross_attention_maps}.
We follow the original VAR inference implementation and apply Gumbel softmax sampling~\cite{gumbel_softmax} with a decreasing temperature, starting from the fifth scale.
At the first four scales, we use nucleus sampling with top-$k{=}400$ and top-$p{=}0.95$. 

\paragraph{Results.}
We provide the results of a side-by-side comparison of \ourmethod against the baselines for various image quality aspects in~\Cref{fig:sbs_baselines}.
As follows from the human evaluation, \ourmethod outperforms all AR baselines in most aspects, only lagging behind Lumina-mGPT, Emu3 and LlamaGen in terms of image complexity.
As for diffusion models, DMD2 slightly outperforms \ourmethod in terms of defects presence, which can be attributed to its adversarial training, SD3-medium is better at text alignment, likely due to an additional text encoder, whereas SDXL surpasses our model in aesthetics.
In other comparisons, \ourmethod is on par with the diffusion models and their distilled versions, with respect to statistical error.
We provide qualitative comparisons in~\Cref{fig:model_comparison_2,fig:model_comparison_1}.

\subsection{Inference performance evaluation}
Next, we analyze the efficiency of \ourmethod's sampling and compare it to the baselines.
We consider two settings: measurement of a single generator step without considering time for text encoding and VAE decoding; and inference time of a full text-to-image pipeline. 
All models are evaluated in half-precision with a batch size of $8$.
KV-cache is enabled for all AR models.
All models are evaluated on a single NVIDIA A100 80GB GPU.
\begin{table}[h]
\small
\centering
\setlength\tabcolsep{3.5pt}
\begin{tabular}{lcccc}
\toprule
Model & \begin{tabular}{@{}c@{}}Generator\\size, B\end{tabular} & N steps & \begin{tabular}{@{}c@{}}1 step,\\ms/image\end{tabular} & \begin{tabular}{@{}c@{}}Full,\\s/image\end{tabular} \\
\midrule
\multicolumn{5}{c}{Diffusion Models} \\
\midrule
SDXL-Turbo & 2.6 & 4 & 42.1 & 0.4 \\
DMD2 & 2.6 & 4 & 42.1 & 0.4 \\
SDXL & 2.6 & 25 & 42.1 & 2.3 \\
SD3-medium & 2.0 & 28 & 61.3 & 3.9 \\
\midrule
\multicolumn{5}{c}{Autoregressive Models} \\
\midrule
Lumina-mGPT & 7.0 & 1024 & ---  & 527.2 \\
LlamaGen & 0.8 & 1024 & ---  & 3.8\textsuperscript{*} \\
HART & 0.7 & 14 & 4.9\textsuperscript{**} & 0.5 \\
\ourmethod (AR) & 2.5 & 14 & 33.2\textsuperscript{**} & 0.6 \\
\midrule
\ourmethod & 2.5 & 14 & 26.4\textsuperscript{**} & 0.5 \\
\bottomrule
\multicolumn{5}{l}{\textsuperscript{*}Can only generate in 512x512.} \\
\multicolumn{5}{l}{\textsuperscript{**}Time averaged over 14 steps.}
\end{tabular}

\vspace{-4mm}
\caption{Comparison of $1024{\times}1024$ image generation times.
}
\vspace{-9pt}
\label{tab:inference_time}
\end{table}

As follows from~\Cref{tab:inference_time}, \ourmethod is among the most efficient image generation models of similar size, being more than 4 times faster than SDXL.
Notably, \ourmethod takes the same time as HART to generate a batch of $1024{\times}1024$ images while being more than three times larger.
This efficiency stems from the fact that we do not employ an additional diffusion model during de-tokenization in RQ-VAE, from the transitioning to a non-causal architecture, and from disabling classifier-free guidance at the latest scales.
\Cref{tab:inference_time_512} presents the evaluation on $512{\times}512$ resolution.

\vspace{-2pt}
\subsection{Ablation study}
\label{sect:ablation}
\begin{figure}
\centering
\includegraphics[width=\linewidth]{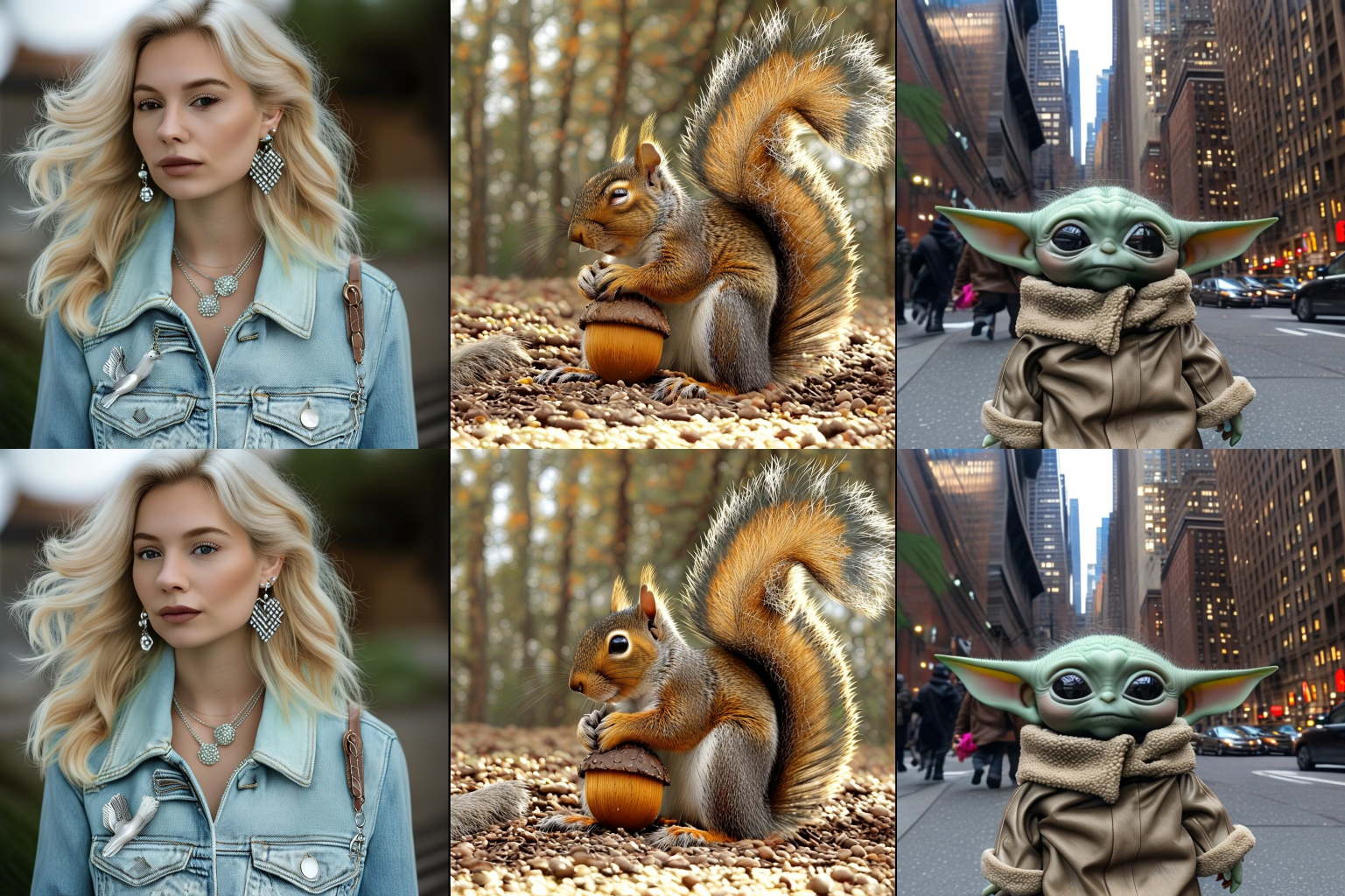}
\vspace{-15pt}
\caption{Illustrative examples when disabled CFG at the last scales (Bottom) mitigates the artifacts in fine-grained details (Top).}
\label{fig:disable_cfg}
\end{figure}

Finally, we evaluate the effect of our architectural choices on the image generation quality and inference time.
The quantitative evaluation in~\Cref{tab:tvars_sbs_comparison,tab:switti_1024_metrics} demonstrates that \ourmethod is not only more sample efficient than its AR alternative, but also exhibits slightly better visual quality with respect to all aspects of human evaluation and automated metrics. 
Moreover, disabling classifier-free guidance at the last two scales noticeably reduces defect presence in \ourmethod's images without affecting other aspects of evaluation, as illustrated in~\Cref{fig:disable_cfg}.
Nevertheless, it should be noted that enabling CFG at the last scales can still be beneficial.
For example, in scenarios where prompts contain text to be rendered in a small font size, guidance can improve the spelling.

\begin{table}[ht]
\scriptsize
\setlength\tabcolsep{3.5pt}
\centering
\begin{tabular}{cccccc}
\toprule
Setup 1 & Setup 2 &  Relevance $\uparrow$ & Aesthetics $\uparrow$ & Complexity $\uparrow$ & Defects $\uparrow$ \\
\midrule
\ourmethod & \scriptsize{\ourmethod AR} & \textcolor{lightgreen}{0.55\textsubscript{0.06}} & \textcolor{lightgreen}{0.55\textsubscript{0.06}} &
\textcolor{lightgreen}{0.52\textsubscript{0.06}} & \textcolor{stronggreen}{0.60\textsubscript{0.06}} \\
\midrule
\ourmethod &  \begin{tabular}{@{}c@{}}\scriptsize{\ourmethod}\\\scriptsize{+ late CFG}\end{tabular} & \textcolor{gray}{0.50\textsubscript{0.06}} & \textcolor{gray}{0.50\textsubscript{0.06}} & \textcolor{gray}{0.51\textsubscript{0.06}} &
\textcolor{stronggreen}{0.56\textsubscript{0.06}} \\
\bottomrule
\end{tabular}

\vspace{-4mm}
\caption{Human evaluation of \ourmethod design choices.
Scores are the mean of Setup 1 wins plus half-ties, with subscripts denoting half the 95\% confidence interval. 
}
\vspace{-7pt}
\label{tab:tvars_sbs_comparison}
\end{table}

\begin{table}[b]
\vspace{-5pt}
\small
\setlength\tabcolsep{2pt}
    \centering
    \scalebox{0.75}{
    \begin{tabular}{lcccc|cccc|c}
    \toprule
    & \multicolumn{4}{c}{COCO} & \multicolumn{4}{c}{MJHQ} & \\
    \midrule
    Setup & PS $\uparrow$ & CLIP $\uparrow$ & IR $\uparrow$ & FID $\downarrow$ & PS $\uparrow$ & CLIP $\uparrow$ & IR $\uparrow$ & FID $\downarrow$ & GenEval \\
\midrule
         \ourmethod (AR) & 0.227 & 0.351 & 0.91 & 18.7 & 0.217 & 0.375 & 0.76 & 8.3 & 0.62 \\
         \ourmethod w/ CFG & 0.228 & 0.353 & 0.90 & \textbf{17.8} & 0.218 & 0.375 & 0.75 & \textbf{8.1} & 0.62 \\
         \ourmethod & \textbf{0.229} & \textbf{0.355} & \textbf{0.96} & 18.2 & \textbf{0.219} & \textbf{0.378} & \textbf{0.81} & \textbf{8.1} & 0.62 \\
         \bottomrule
    \end{tabular}
    }
    \vspace{-4pt}
    \caption{Impact of the acceleration modifications at $1024{\times}1024$.}
    \label{tab:switti_1024_metrics}
\end{table}

As we show in~\Cref{tab:tvars_speedup_comparison}, a non-autoregressive architecture improves sampling efficiency, reducing latency by around $21\%$.
Disabling CFG on the last two scales further decreases latency by an additional $32\%$.
Moreover, by removing the causality in \ourmethod, we free up to $2.3$ GB of GPU memory (in half precision), required to store the KV caches of the self-attention layers during the sampling for a single image.

\begin{table}[]
\small
\centering
\setlength\tabcolsep{2.75pt}
\begin{tabular}{lc|ccc}
\toprule
Model & \begin{tabular}{@{}c@{}}KV cache,\\ GB/image\end{tabular} & \begin{tabular}{@{}c@{}}Generator step,\\ms/image\end{tabular} & \begin{tabular}{@{}c@{}}Disable\\ late CFG\end{tabular} & \begin{tabular}{@{}c@{}}Full,\\ms/image\end{tabular} \\
\midrule
\multirow{2}{*}{\ourmethod (AR)} & 2.3 & \multirow{2}{*}{465} & \ding{55} & 908 \\
 & 1.2 & & \checkmark & 640 \\
\multirow{2}{*}{\ourmethod} & 0 & \multirow{2}{*}{369} & \ding{55} & 735 \\
& 0 & & \checkmark & 540 \\
\bottomrule
\end{tabular}

\vspace{-7.5pt}
\caption{Efficiency comparison of architecture and sampling modifications proposed in \ourmethod.}
\label{tab:tvars_speedup_comparison}
\vspace{-7.5pt}
\end{table}

\section{Limitations and future directions}

\paragraph{Hierarchical tokenizers.}
We believe that one of the major limitations of the existing scale-wise models is the inferior hierarchical discrete VAE performance compared to the recent continuous~\cite{chen2024deepcompression, cosmos_tokenizer, podell2024sdxl, flux} or discrete single-level~\cite{cosmos_tokenizer, sun2024autoregressive} counterparts. 

Typical failure cases are distorted middle/long-shot faces, text rendering and checker-board artifacts on high-frequency textures such as distant foliage or rocky surfaces.  
We hope that future advances in hierarchical image tokenizers, either discrete or continuous, may significantly improve the performance of scale-wise generative models without using an additional diffusion prior~\cite{tang2024hart, li2024mar}.

\section{Conclusion}

We introduce \ourmethod, a novel scale-wise generative transformer for text-to-image generation. 
In contrast to previous next-scale prediction approaches, \ourmethod eliminates explicit autoregressive prior and makes use of more effective sampling with guidance.
These modifications result in \ourmethod generating images as fast as HART, while being three times larger.
Trained on a large-scale curated text-image dataset, \ourmethod surpasses prior text-conditional visual autoregressive models and achieves up to $7{\times}$ faster sampling than state-of-the-art text-to-image diffusion models while delivering comparable generation quality.

{
    \small
    \bibliographystyle{ieeenat_fullname}
    \bibliography{main}
}

\appendix
\clearpage
\setcounter{page}{1}
\maketitlesupplementary

\section{Training details}
\label{app:training_details_analysis}
For all our experiments, we use FSDP with a hybrid strategy for effective multi-host training and mixed precision BF16/FP32.  
To additionally reduce memory usage and speed up the training steps, we use precomputed textual embeddings from the text encoders.
We use AdamW optimizer with ($\beta_1{=}0.9$, $\beta_2{=}0.95$).
In the normalized RoPE, we use $\theta{=}10,000$ and max size $128$.

\subsection{Pretraining}

\paragraph{Data.}
Starting from ${\sim}6$B image-text pairs from the web, we filter out images of low aesthetic quality, based on the AADB~\cite{kong2016photo} and TAD66k~\cite{he2022rethinking} aesthetic filters. 
We additionally consider sufficiently high-resolution images with at least $512$px on each side.
The resulting dataset contains central crops of the images with an aspect ratio in $[0.75, 1.33]$. 
The images are recaptioned using the LLaVA-v1.4-13B, LLaVA-v1.6-34B~\cite{liu2023llava}, and ShareGPT4V~\cite{chen2023sharegpt4v} models. 
The best caption is selected according to OpenCLIP ViT-G/14~\cite{clip_laion}.

\paragraph{Technical details.}
During the first stage of pretraining, we train $d{=}30$ models on $256{\times}256$ resolution for $400$K iterations using a batch size of $2,560$.
This stage takes ${\sim}25$K NVIDIA A100 GPU hours.
We start with a learning rate of $1e{-}4$ with a linear decay to $1e{-}5$.

Next, we train the models on $512{\times}512$ resolution for $200$K iterations using a batch size of $768$ and a learning rate of $1e{-}5$, linearly decaying to $5e{-}7$.
This stage takes another ${\sim}12$K NVIDIA A100 GPU hours.

\subsection{Ablation experiments}
For the experiments in~\Cref{sec:our_architecture}, we train more light-weight models, with $d{=}20$ transformer blocks, resulting in approximately $0.7$B parameters.
All models are trained for $150,000$ iterations with a learning rate of $1e{-}4$, linearly decaying to $1e{-}5$. 
Batch size is $768$.
Image resolution is $256{\times}256$.
For these experiments, we disable the conditioning on the cropping parameters. 

For evaluation, we use $30,000$ prompts from the COCO2014 validation set~\cite{ms_coco}.

\section{Normalization ablation}
\label{app:normalizations}
In this section, we investigate, how the choice of normalization affects the final model performance.
In~\Cref{tab:d20_norm_ablation}, we compare our $d{=}20$ version of \ourmethodar, trained on $256{\times}256$ against the same version with switched normalization functions (i.e RMSNorms are used for QK-normalization in attention, and LN for input/output normalization) and against the version without ``sandwich'' normalization.
We find the difference in models' performance insignificant. 

\begin{table}[h]
    \centering
    \begin{tabular}{lccc}
    \toprule
    Model ($d{=}20$) & PickScore $\uparrow$ & CLIP Score $\uparrow$ & FID $\downarrow$ \\
    \midrule
    Switti (AR) & 0.206 & 0.312 & 10.8 \\
    LN $\leftrightarrow$ RMSNorm & 0.206 & 0.313 & 11.1 \\
    w/o ``sandwich'' & 0.206 & 0.311 & 11.0 \\
    \bottomrule
    \end{tabular}
    \caption{Ablation of normalization choices for $d{=}20$ models trained on $256{\times}256$ images.}
    \label{tab:d20_norm_ablation}
\end{table}

\section{Training loss analysis}
\label{app:non_causal_loss}

\begin{figure}[h]
    \centering
    \includegraphics[width=\linewidth]{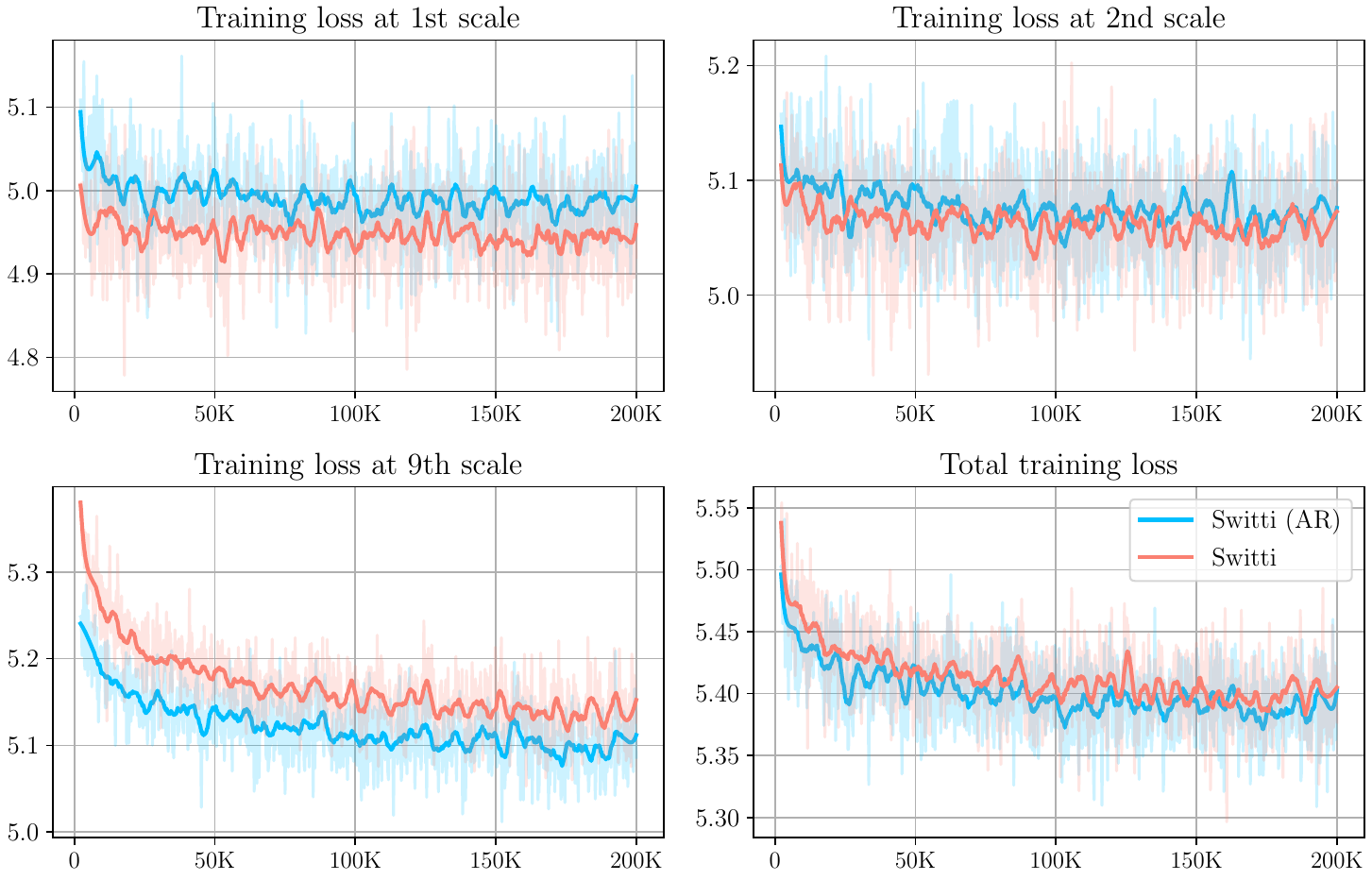}
    \caption{Comparison of training losses at various scales on $512{\times}512$ resolution for $d{=}30$ models.}
    \label{fig:loss_comparison}
\end{figure}

We find that both models converge to similar total losses at the end of the training, as illustrated in~\Cref{fig:loss_comparison}. 
However, \ourmethodar has slightly lower values of loss at later scales, while the non-AR version exhibits marginally better performance at the earlier ones.
This pattern holds for all configurations of our model during all stages of the training.

This phenomenon combined with our observation in~\Cref{sec:sft} that around 80\% of image tokens can be replaced without affecting the final image visual quality, and the fact that the final non-causal model slightly outperforms its AR counterpart, indicates that the accuracy at the last scales is less crucial.

\section{Additional prompt switching visualizations}
\label{app:switch_prompt}

\begin{figure*}[]
\centering
\includegraphics[width=\linewidth]{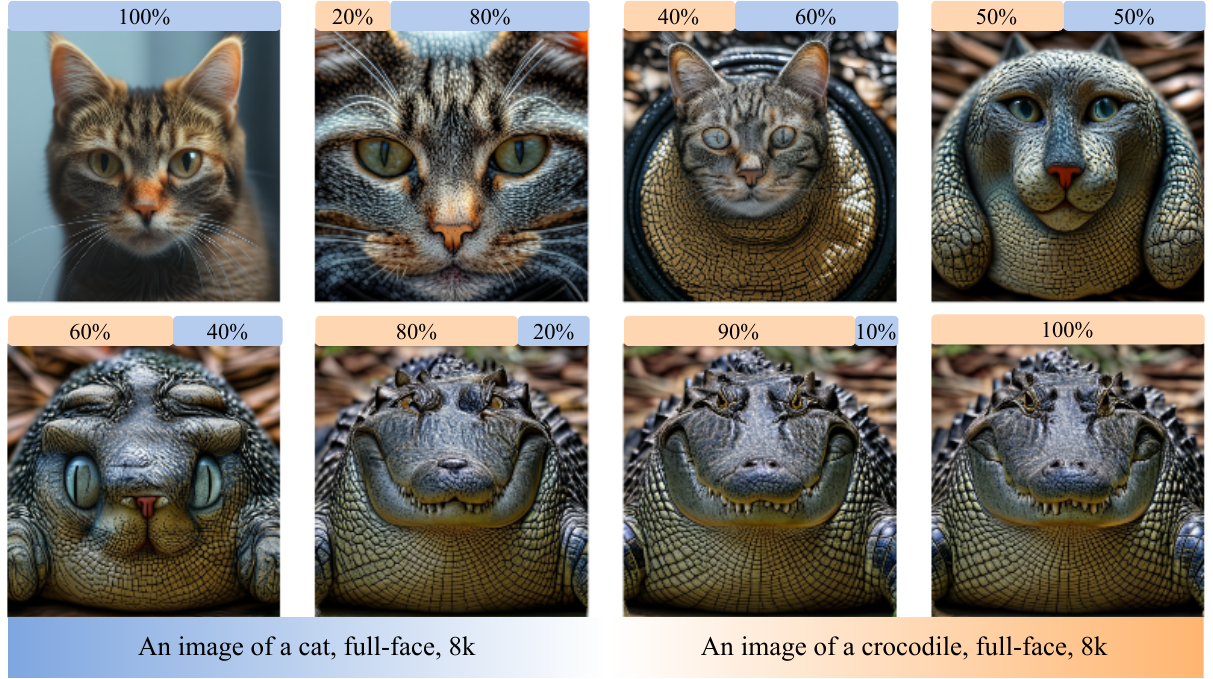}
\caption{Impact of the prompt switching during image generation.}
\label{fig:switch_prompt_app-1}
\end{figure*}

\begin{figure*}[]
\centering
\includegraphics[width=\linewidth]{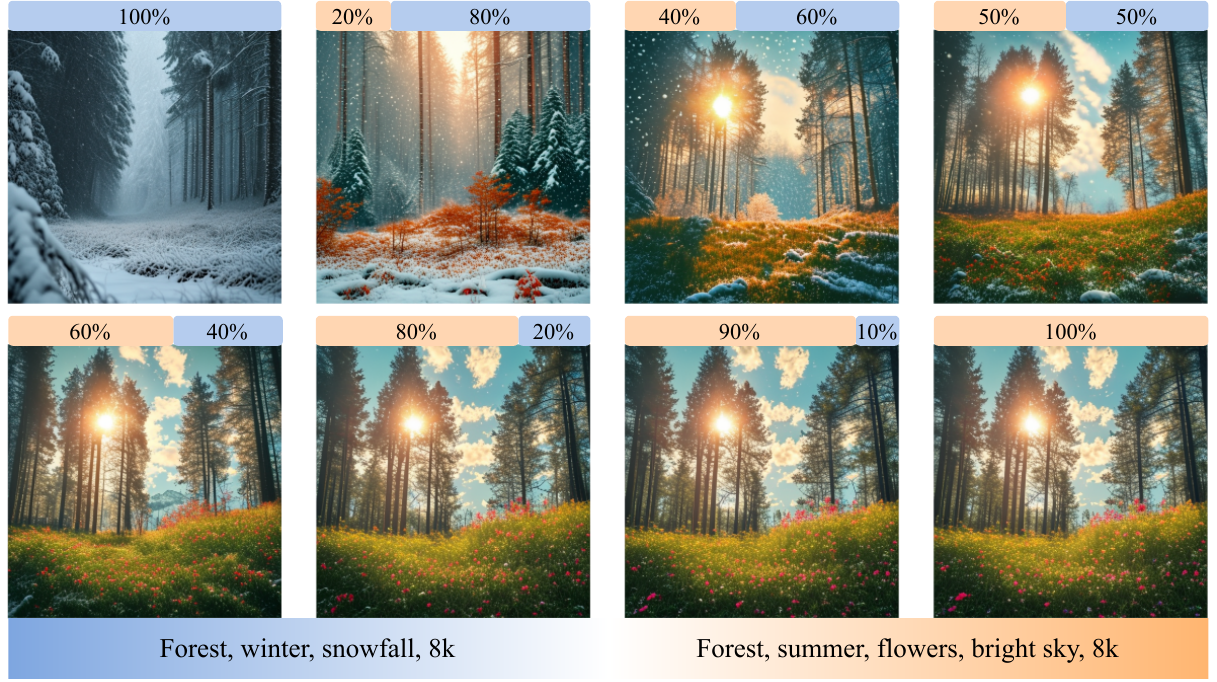}
\caption{Impact of the prompt switching during image generation.}
\label{fig:switch_prompt_app-2}
\end{figure*}

In~\Cref{fig:switch_prompt_app-1,fig:switch_prompt_app-2}, we provide additional examples of the prompt switching analysis, discussed in~\Cref{sec:cross_attention_maps}.

\section{Visual comparison of the finetuned RQ-VAE}
\label{app:vae_visual_comparison}

To illustrate the difference between the original RQ-VAE checkpoint and fine-tuned version, we depict several representative examples in Figure~\ref{fig:vae_visual_comparison}. 
One can observe that the fine-tuned
VAE decoder is less prone to reconstruction artifacts and color shifts and produces more contrast images. 

\section{Evaluation details}
\label{app:evaluation details}
We compute the CLIP Score~\cite{clip_score}, using features from a pre-trained CLIP-ViT-H-14-laion2B-s32B-b79K encoder~\cite{clip_laion}, to assess image-text alignment.

FID is measured on $30$K generated images reduced to $512{\times}512$ resolution using bicubic interpolation. 
Then, the resolution is further reduced to $256{\times}256$ using Lanczos interpolation following the practices in FID calculation on COCO2014.
Real data statistics are collected for all images in the validation sets: ${\sim}40$K images in COCO2014 and $30,000$ images in MJHQ.

For GenEval, we generate 4 images for each of the 533 evaluation prompts, followed by the original evaluation protocol using Mask2Former~\cite{mask2former} with Swin-S backbone as an object detector.
\section{Human evaluation setup}
\label{app:human_eval}

The evaluation is performed using Side-by-Side (SbS) comparisons, i.e., the assessors are asked to make a decision between two images given a textual prompt.
For each evaluated pair, three responses are collected and the final prediction is determined by majority voting.

The human evaluation is performed by professional assessors. 
They are officially hired, paid competitive salaries, and informed about potential risks.
The assessors have received detailed and fine-grained instructions for each evaluation aspect and passed training and testing before accessing the main tasks.

\begin{figure}[htb]
\begin{subfigure}{\linewidth}
\centering
\includegraphics[width=\linewidth]{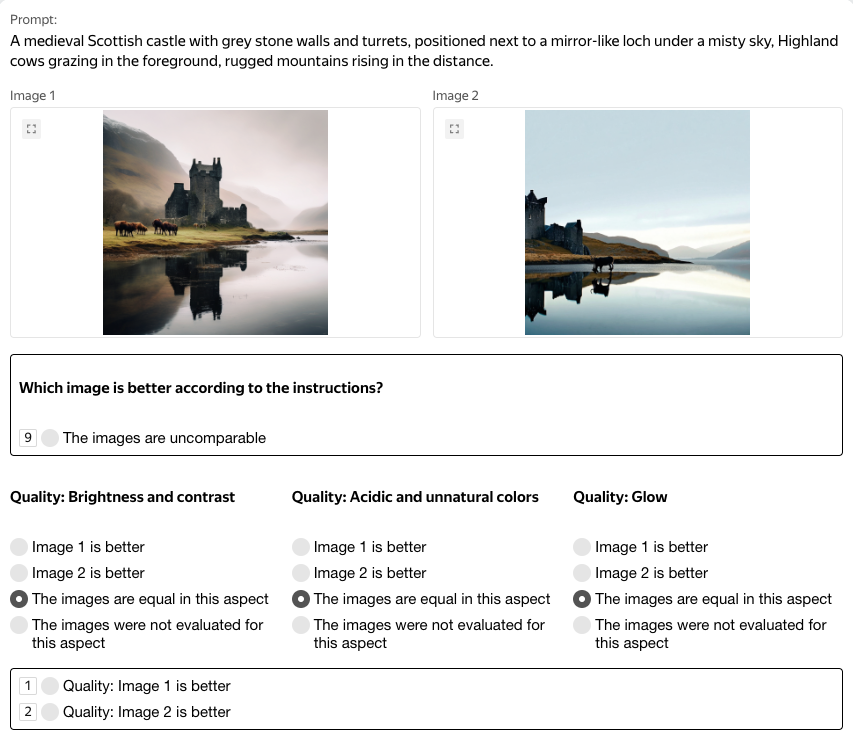}
\end{subfigure}
\begin{subfigure}{\linewidth}
\centering
\includegraphics[width=\linewidth]{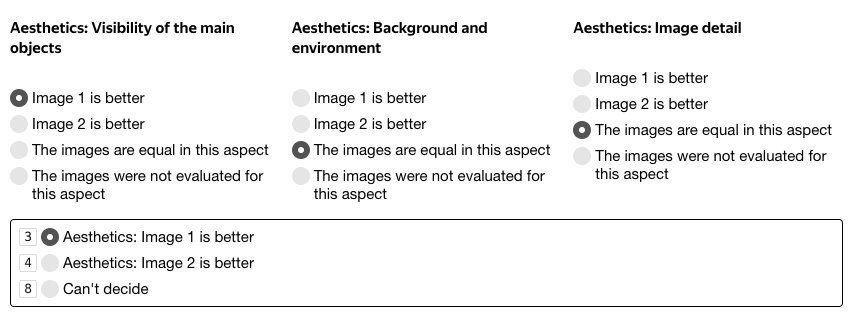}
\end{subfigure}
\caption{Human evaluation interface for \textbf{aesthetics}.}
\label{fig:human_eval_aest}
\end{figure}

\begin{figure}[htb]
\begin{subfigure}{\linewidth}
\centering
\includegraphics[width=\linewidth]{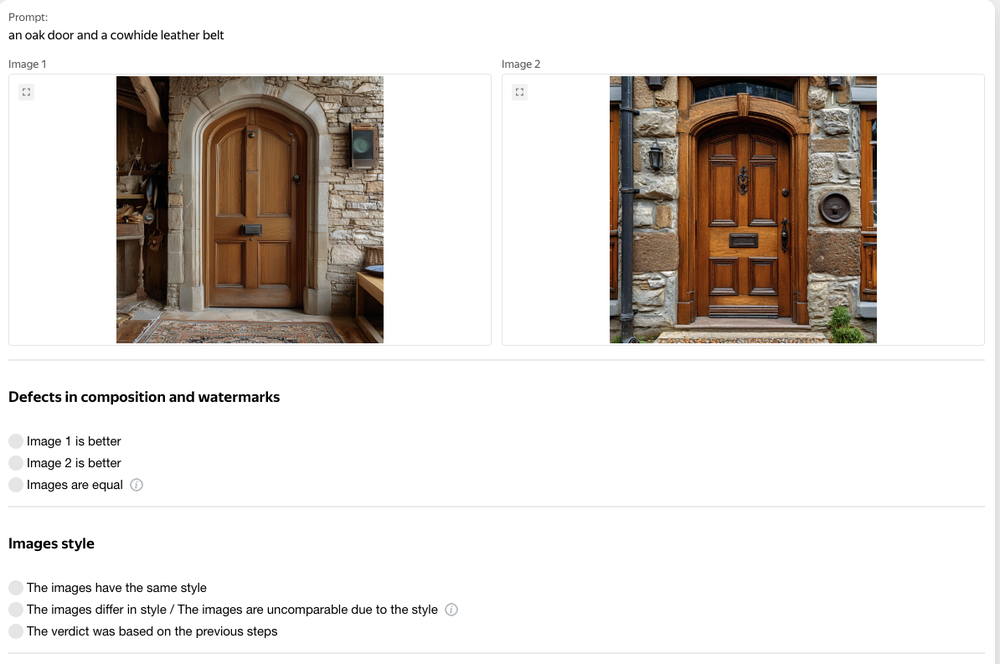}
\end{subfigure}
\begin{subfigure}{\linewidth}
\centering
\includegraphics[width=\linewidth]{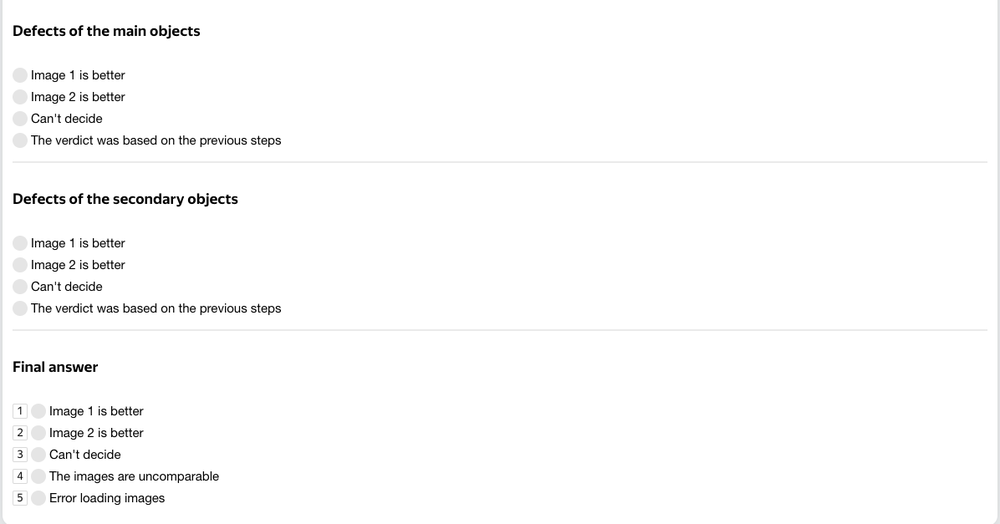}
\end{subfigure}
\caption{Human evaluation interface for \textbf{defects}.}
\label{fig:human_eval_defect}
\end{figure}

\begin{figure}[htb]
\begin{subfigure}{\linewidth}
\centering
\includegraphics[width=\linewidth]{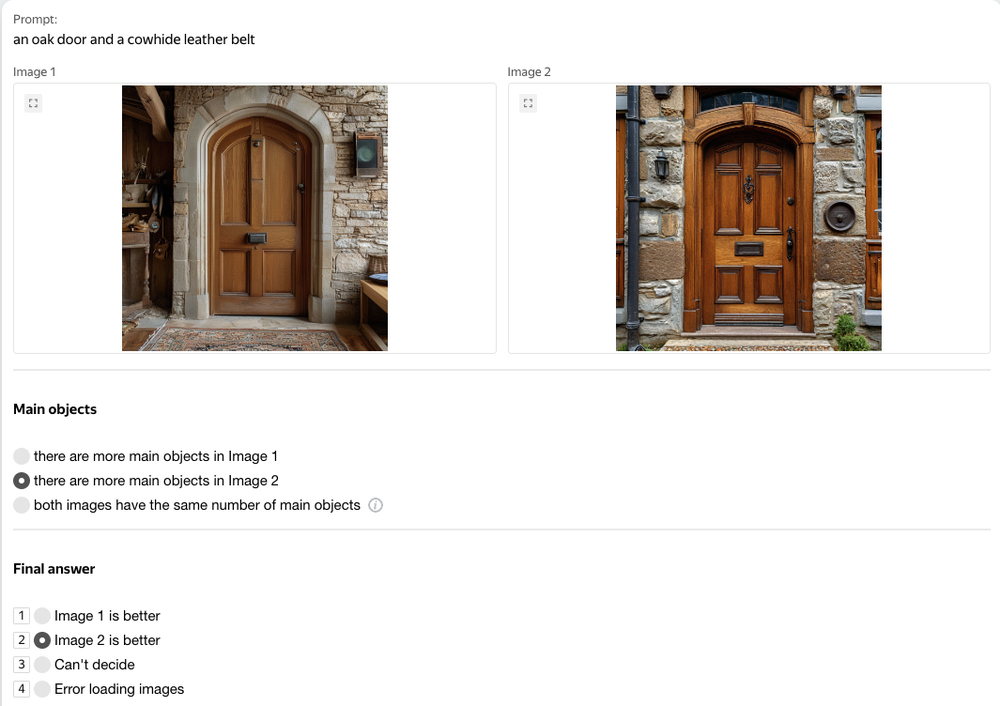}
\end{subfigure}
\begin{subfigure}{\linewidth}
\centering
\includegraphics[width=\linewidth]{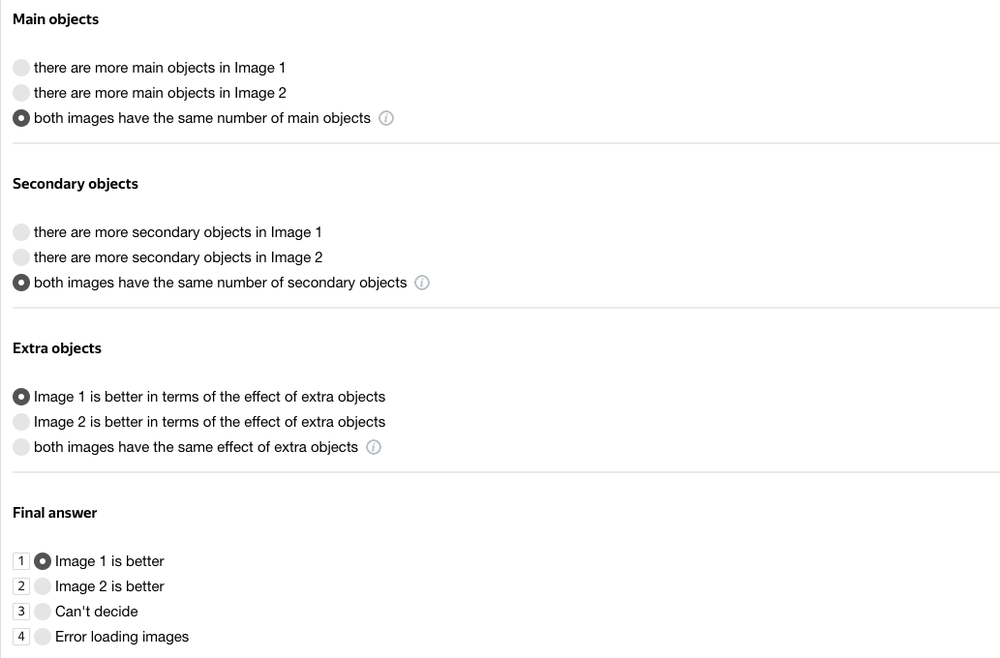}
\end{subfigure}
\caption{Human evaluation interface for \textbf{relevance}.}
\label{fig:human_eval_rel}
\end{figure}

\begin{figure}[htb]
\centering
\includegraphics[width=\linewidth]{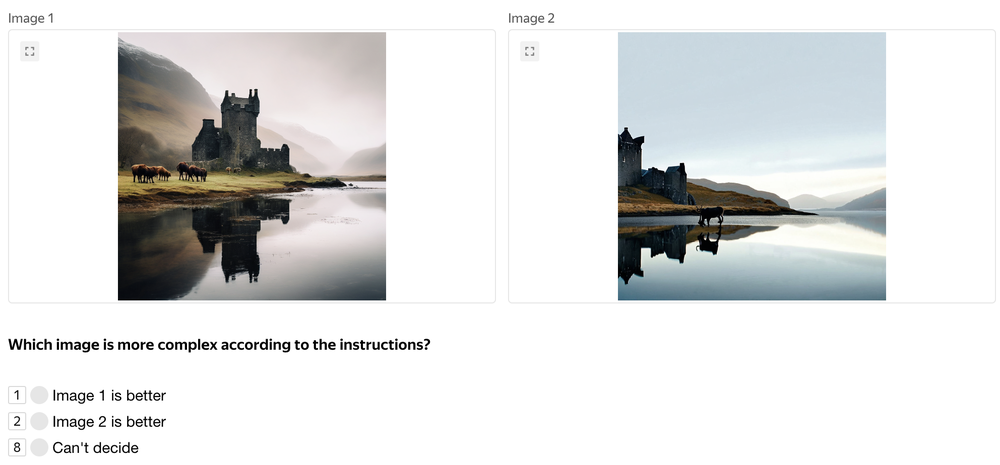}
\caption{Human evaluation interface for \textbf{image complexity}.}
\label{fig:human_eval_compl}
\end{figure}

In our human preference study, we compare the models in terms of four aspects: relevance to a textual prompt, presence of defects, image aesthetics, and complexity.
\Cref{fig:human_eval_aest,fig:human_eval_defect,fig:human_eval_rel,fig:human_eval_compl} present the interface for each of these criteria. 
Note that the selected answers on the images are random.  

\section{Visual comparison against T2I models}
\label{app:visual_comparison}
\begin{figure*}
\centering
\includegraphics[width=\linewidth]{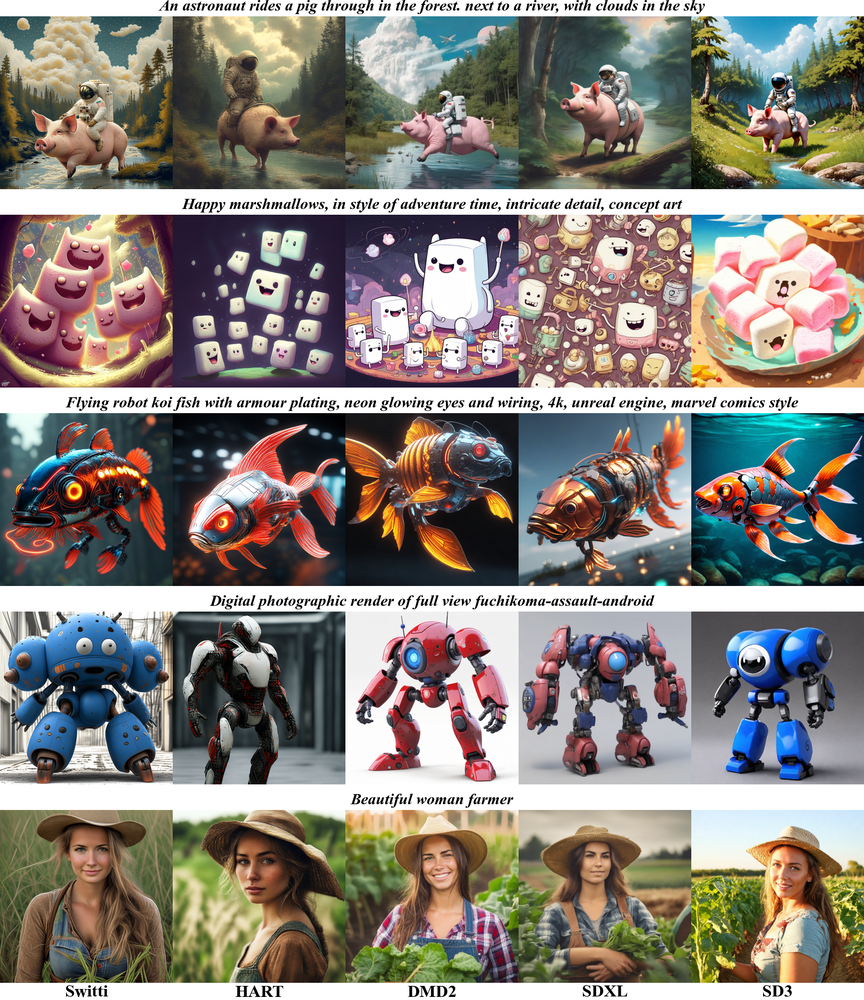}
\caption{Qualitative comparison of \ourmethod against the baselines.}
\label{fig:model_comparison_1}
\end{figure*}

We provide qualitative comparison of \ourmethod against the baselines considered in this work in~\Cref{fig:model_comparison_1} and~\Cref{fig:model_comparison_2}. 

\begin{table}[h]
\small
\centering
\setlength\tabcolsep{3.5pt}
\begin{tabular}{lcccc}
\toprule
Model & \begin{tabular}{@{}c@{}}Generator\\size, B\end{tabular} & N steps & \begin{tabular}{@{}c@{}}1 step,\\ms/image\end{tabular} & \begin{tabular}{@{}c@{}}Full,\\s/image\end{tabular} \\
\midrule
\multicolumn{5}{c}{Distilled Diffusion Models} \\
\midrule
SDXL-Turbo & 2.6 & 4 & 12.4 & 0.25 \\
DMD2 & 2.6 & 4 & 12.4 & 0.25 \\
\midrule
\multicolumn{5}{c}{Diffusion Models} \\
\midrule
SDXL & 2.6 & 25 & 12.4 & 0.87 \\
SD3-medium & 2.0 & 28 & 16.8 & 0.93 \\
\midrule
\multicolumn{5}{c}{Autoregressive Models} \\
\midrule
Lumina-mGPT & 7.0 & 1024 & ---  & 224.2 \\
LlamaGen & 0.8 & 1024 & ---  & 3.82 \\
HART & 0.7 & 10 & 4.7\textsuperscript{*} & 0.06 \\
\ourmethod (AR) & 2.5 & 10 & 11.2\textsuperscript{*} & 0.14 \\
\midrule
\ourmethod & 2.5 & 10 & 9.5\textsuperscript{*} & 0.13 \\
\bottomrule
\multicolumn{5}{l}{\textsuperscript{*}Time averaged over 10 steps.}
\end{tabular}

\caption{Comparison of $512{\times}512$ image generation times.
}
\vspace{-12pt}
\label{tab:inference_time_512}
\end{table}

\section{Effect of disabling CFG at different scales}
\label{app:disable_cfg}
In~\Cref{fig:disable_cfg_2}, we provide some examples of disabling CFG at various level ranges. 
One can observe that presence of CFG at first scales improves image quality and relevance. 
At the same time, it can be turned off at the last scales without noticeable quality degradation or loss of details.

\section{List of prompts used in our figures}

\paragraph{\Cref{fig:main} prompts}

1) \textit{``Cute winter dragon baby, kawaii, Pixar, ultra detailed, glacial background, extremely realistic.``}

2) \textit{``A lizard that looks very much like a man, with developed muscles, leather armor with metal elements, in the hands of a large trident decorated with ancient runes, against the background of a small lake, everything is well drawn in the style of fantasy''}

3) \textit{``An ancient ruined archway on the moon, fantasy, ruins of an alien civilization, concept art, blue sky, reflection in water pool, large white planet rising behind it''}

4) \textit{``Cat as a wizard''}

5) \textit{``The Mandalorian by masamune shirow, fighting stance, in the snow, cinematic lighting, intricate detail, character design''}

6) \textit{``a human face wearing a massive aztec headdress. With hyper intricate war paint on it’s face, detailed texture, vibrant colors, solid background, hyper realistic, 8K``}

7) \textit{``32 – bit pixelated future Hiphop producer in glowing power street ware, noriyoshi ohrai, in the style of minecraft tomer hanuka.``}

8) \textit{``Portrait of an alien family from the 1970’s, futuristic clothes, absurd alien helmet, straight line, surreal, strange, absurd, photorealistic, Hasselblad, Kodak, portra 800, 35mm lens, F 2.8, photo studio.''}

\paragraph{\Cref{fig:disable_cfg} prompts}

1) \textit{``the girl is 36 years old. hairstyle square, blonde hair, slightly curly. light blue jeans. the top is a classic mint-colored jacket. a brooch in the form of a bird is attached to the jacket. there are silver earrings in her ears. a small leather shoulder bag in dark brown color.''}

2) \textit{``a squirrel and an acorn''}

3) \textit{``Baby Yoda Walking in Manhattan.''}

\begin{figure*}[]
\centering
\includegraphics[width=\linewidth]{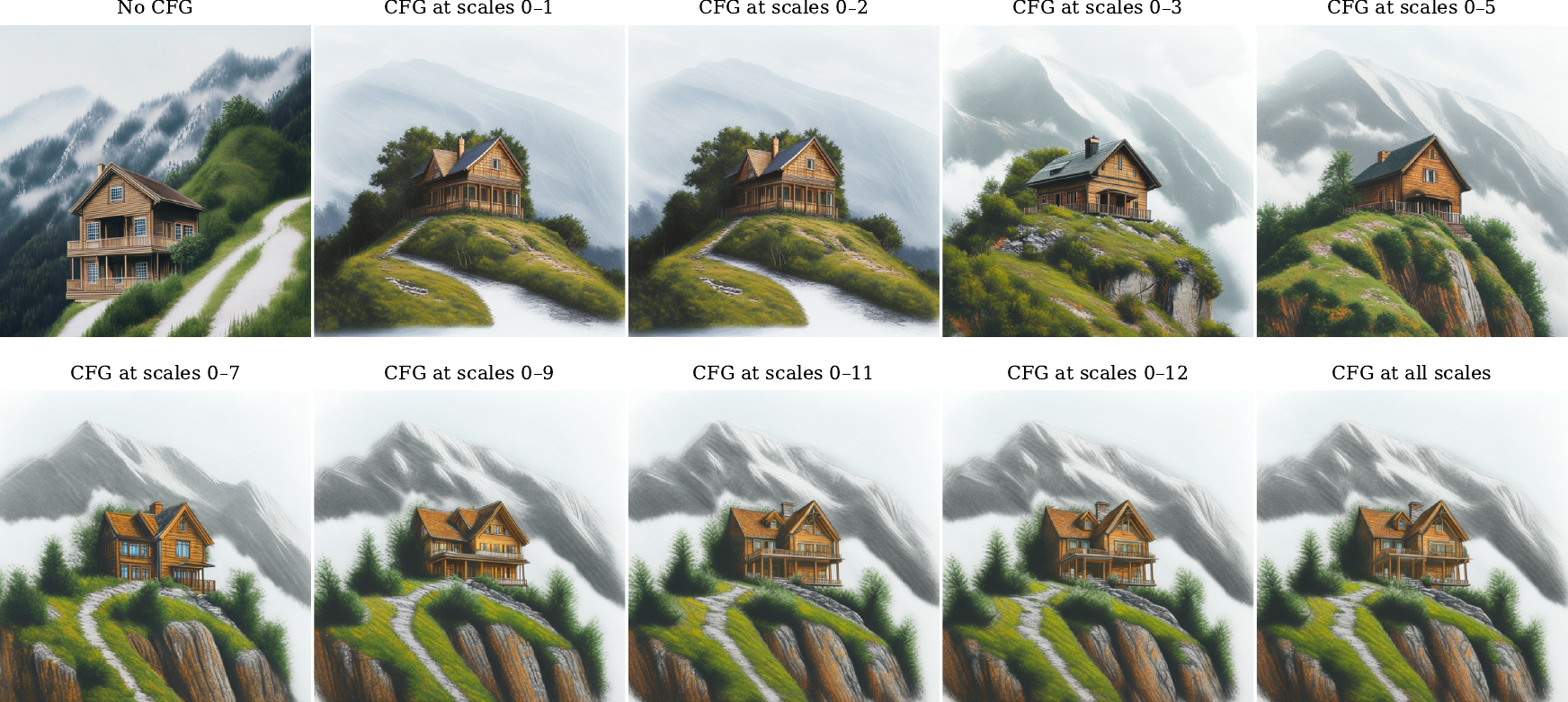}
\includegraphics[width=\linewidth]{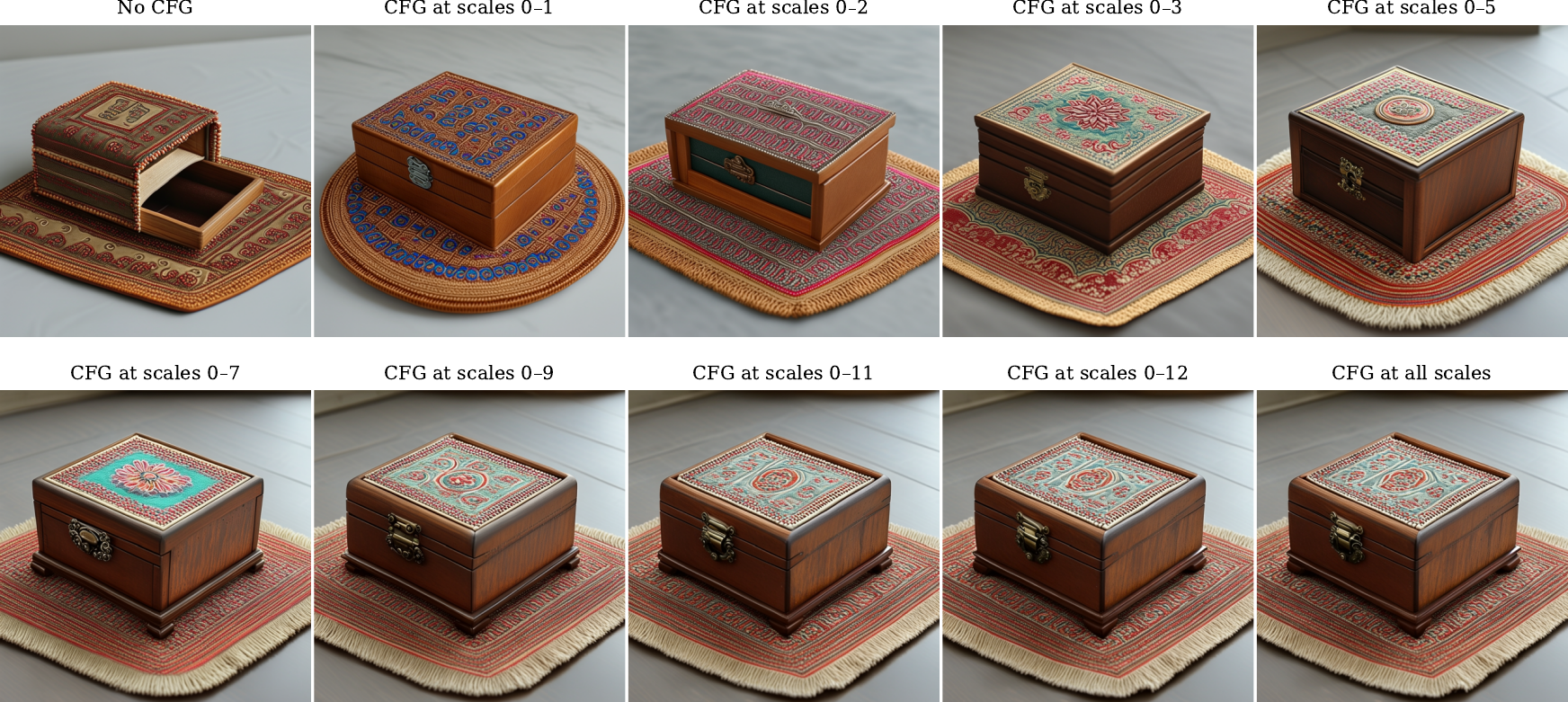}
\caption{Impact of CFG at different resolutions. There are $14$ scales: higher index indicates higher resolution. The guidance scale is $6$.}
\label{fig:disable_cfg_2}
\end{figure*}

\paragraph{\Cref{fig:disable_cfg_2} prompts}

1) \textit{``a drawing of a house on a mountain''}

2) \textit{``a wooden jewelry box and a fabric rug''}

\begin{figure}[htb]
\begin{subfigure}{0.495\linewidth}
\centering
\includegraphics[width=\linewidth]{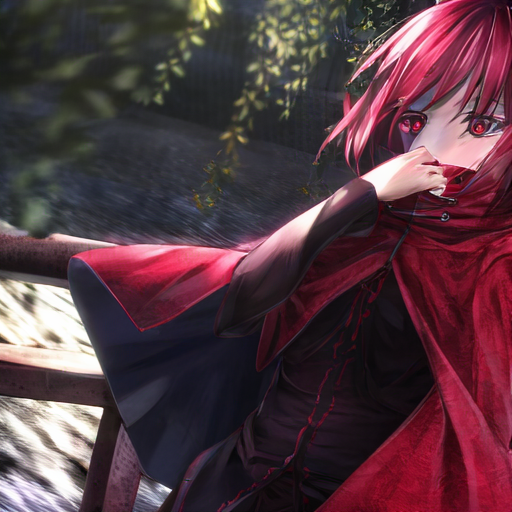}
\end{subfigure}
\hfill
\begin{subfigure}{0.495\linewidth}
\centering
\includegraphics[width=\linewidth]{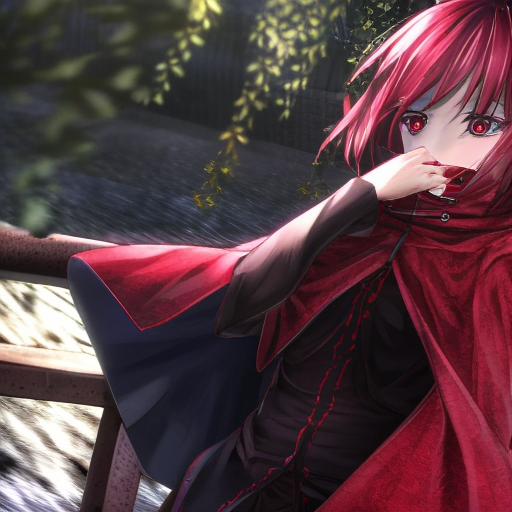}
\end{subfigure}
\begin{subfigure}{0.495\linewidth}
\centering
\includegraphics[width=\linewidth]{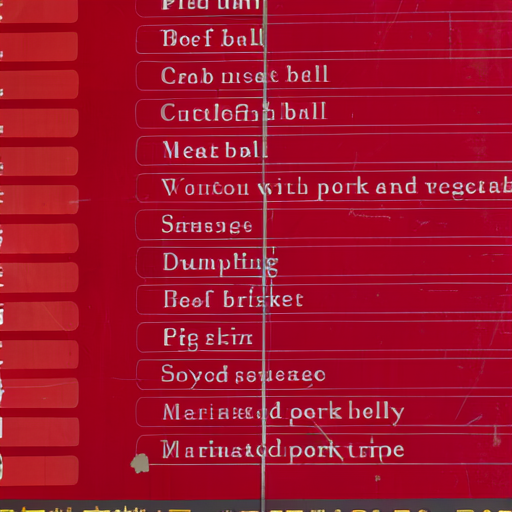}
\end{subfigure}
\hfill
\begin{subfigure}{0.495\linewidth}
\centering
\includegraphics[width=\linewidth]{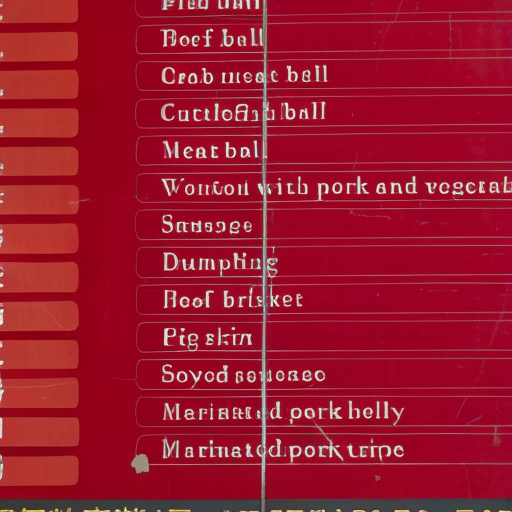}
\end{subfigure}
\begin{subfigure}{0.495\linewidth}
\centering
\includegraphics[width=\linewidth]{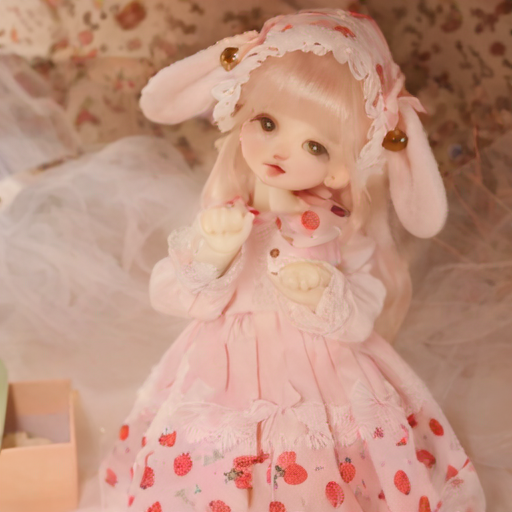}
\end{subfigure}
\hfill
\begin{subfigure}{0.495\linewidth}
\centering
\includegraphics[width=\linewidth]{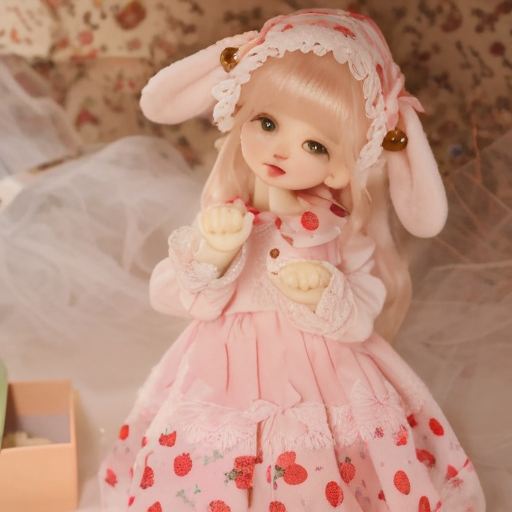}
\end{subfigure}
\begin{subfigure}{0.495\linewidth}
\centering
\includegraphics[width=\linewidth]{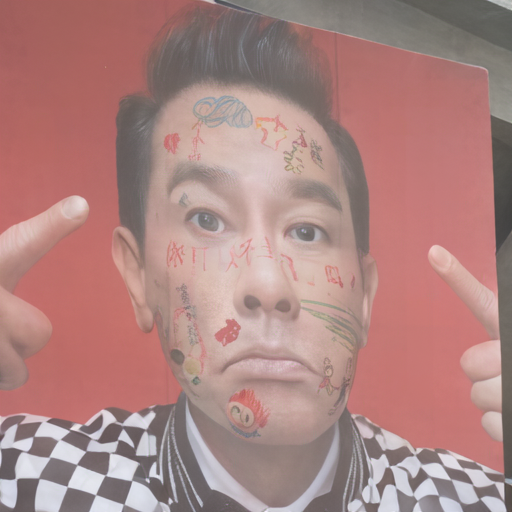}
\end{subfigure}
\hfill
\begin{subfigure}{0.495\linewidth}
\centering
\includegraphics[width=\linewidth]{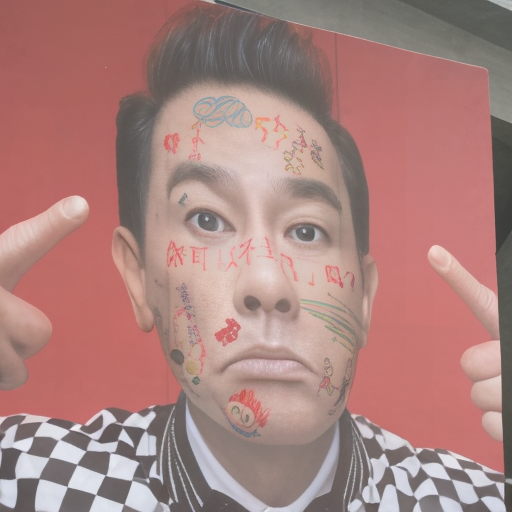}
\end{subfigure}
\caption{Visual comparison of $512{\times}512$ image reconstruction between an original RQ-VAE (\textbf{left}) and the one with a fine-tuned decoder (\textbf{right}).}
\label{fig:vae_visual_comparison}
\end{figure}

\begin{figure}[htb]
\begin{subfigure}{0.495\linewidth}
\centering
\includegraphics[width=\linewidth]{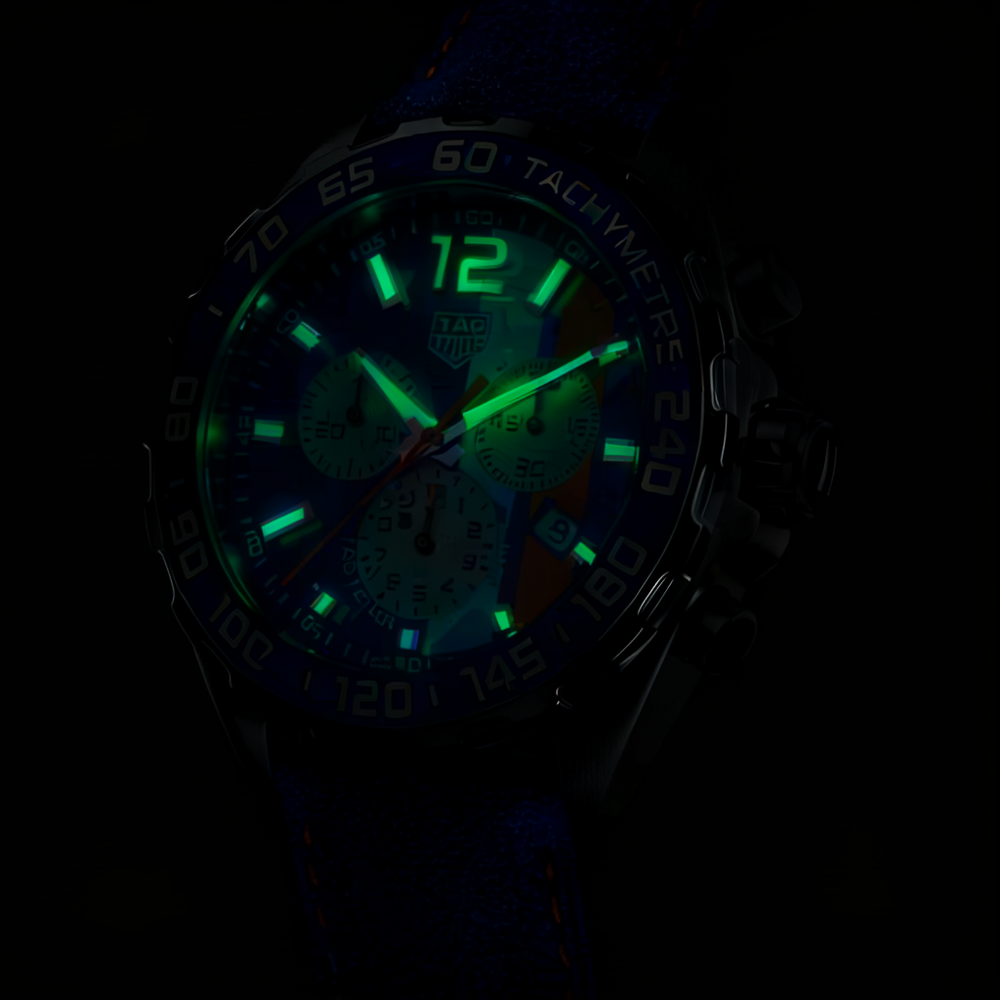}
\end{subfigure}
\hfill
\begin{subfigure}{0.495\linewidth}
\centering
\includegraphics[width=\linewidth]{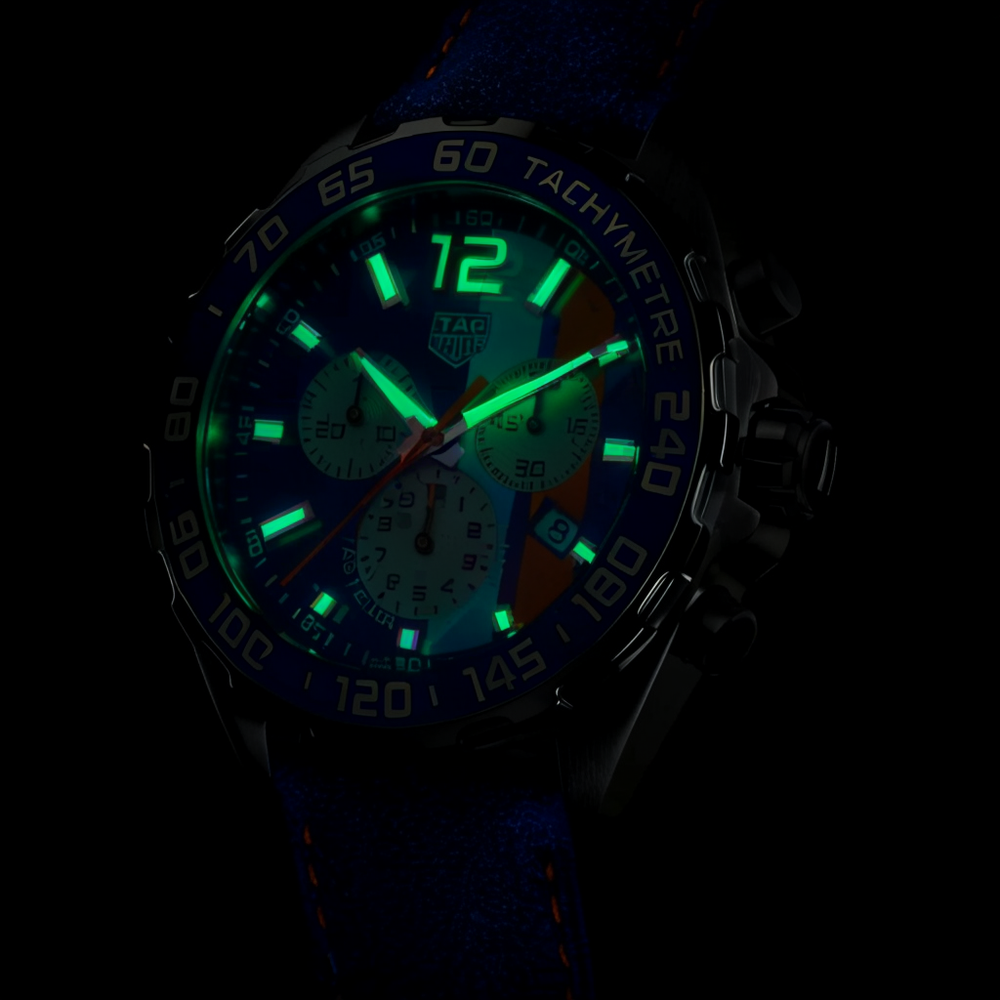}
\end{subfigure}
\begin{subfigure}{0.495\linewidth}
\centering
\includegraphics[width=\linewidth]{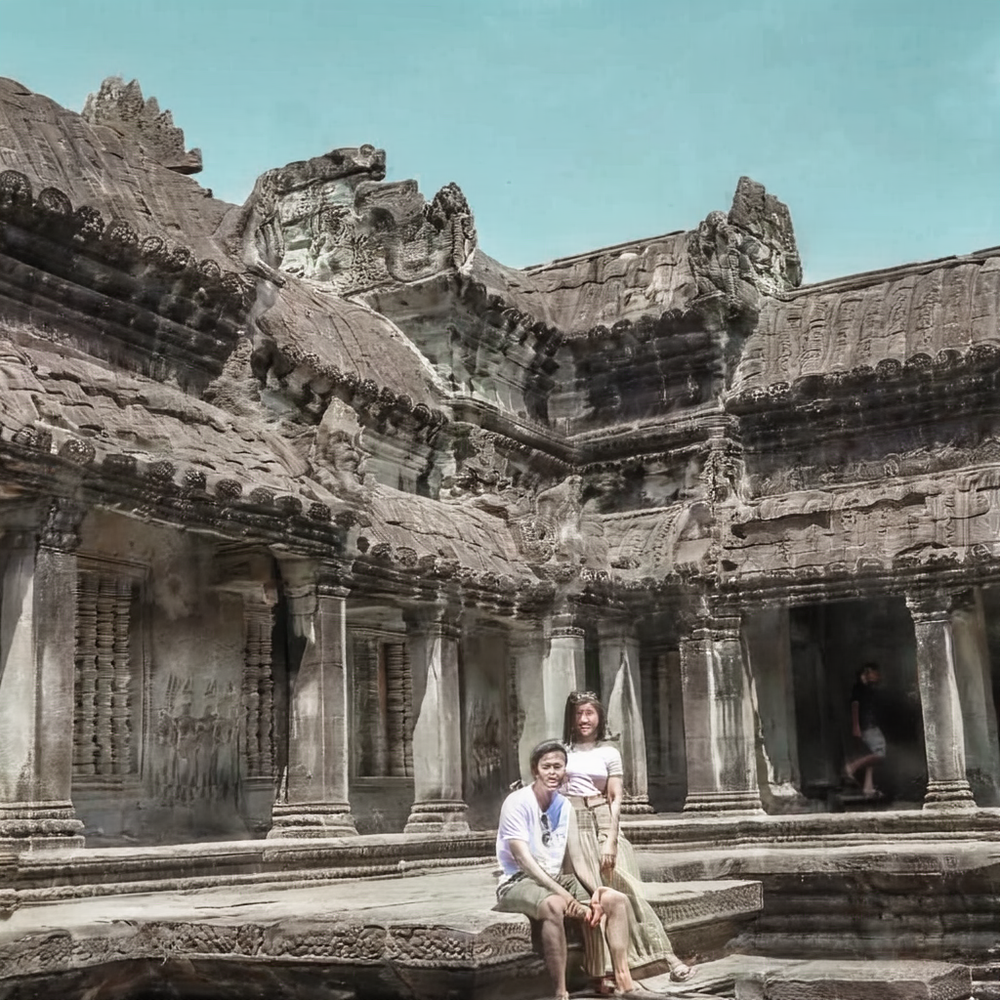}
\end{subfigure}
\hfill
\begin{subfigure}{0.495\linewidth}
\centering
\includegraphics[width=\linewidth]{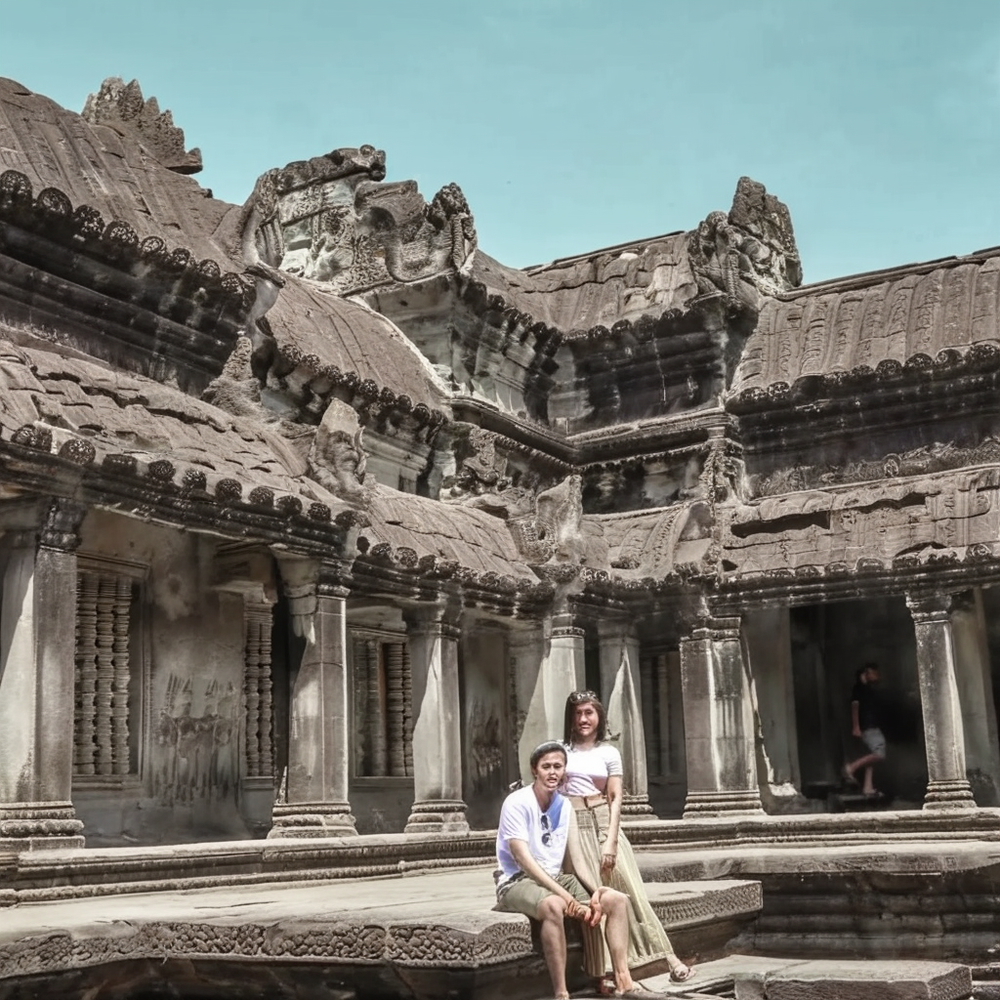}
\end{subfigure}
\begin{subfigure}{0.495\linewidth}
\centering
\includegraphics[width=\linewidth]{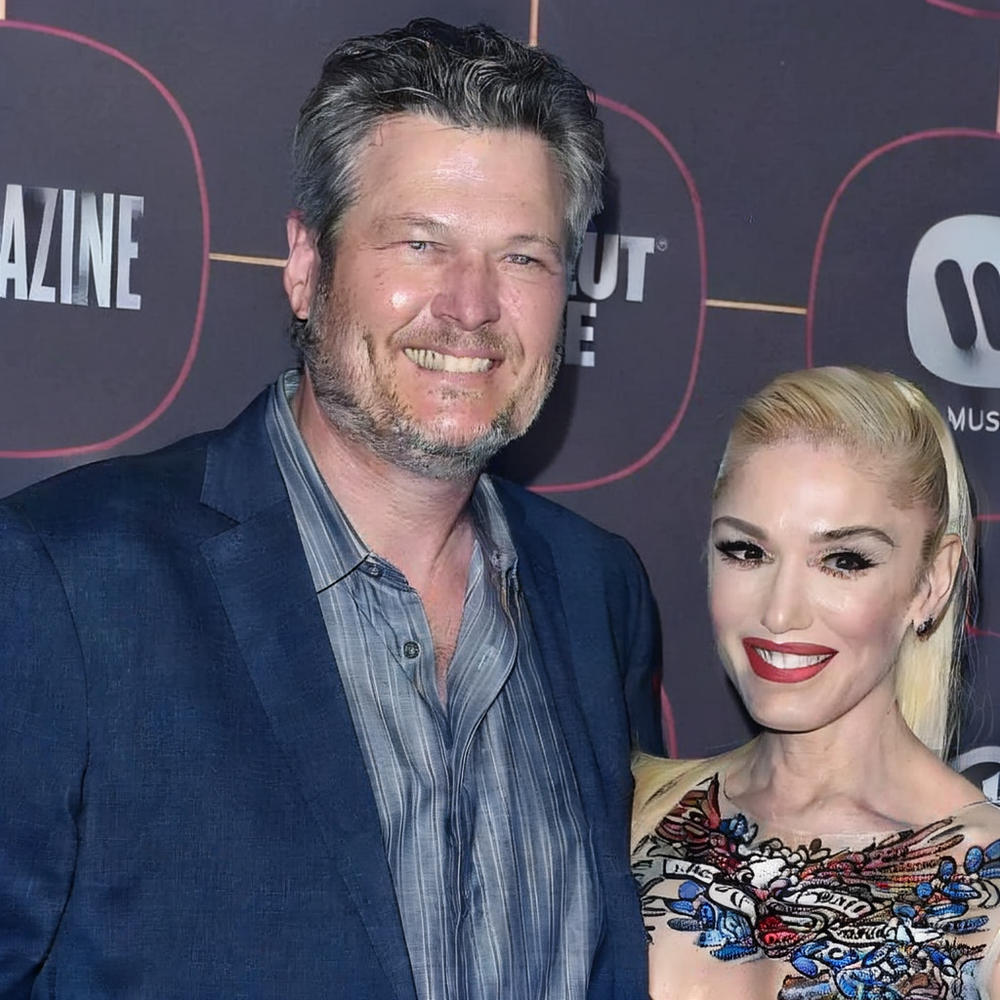}
\end{subfigure}
\hfill
\begin{subfigure}{0.495\linewidth}
\centering
\includegraphics[width=\linewidth]{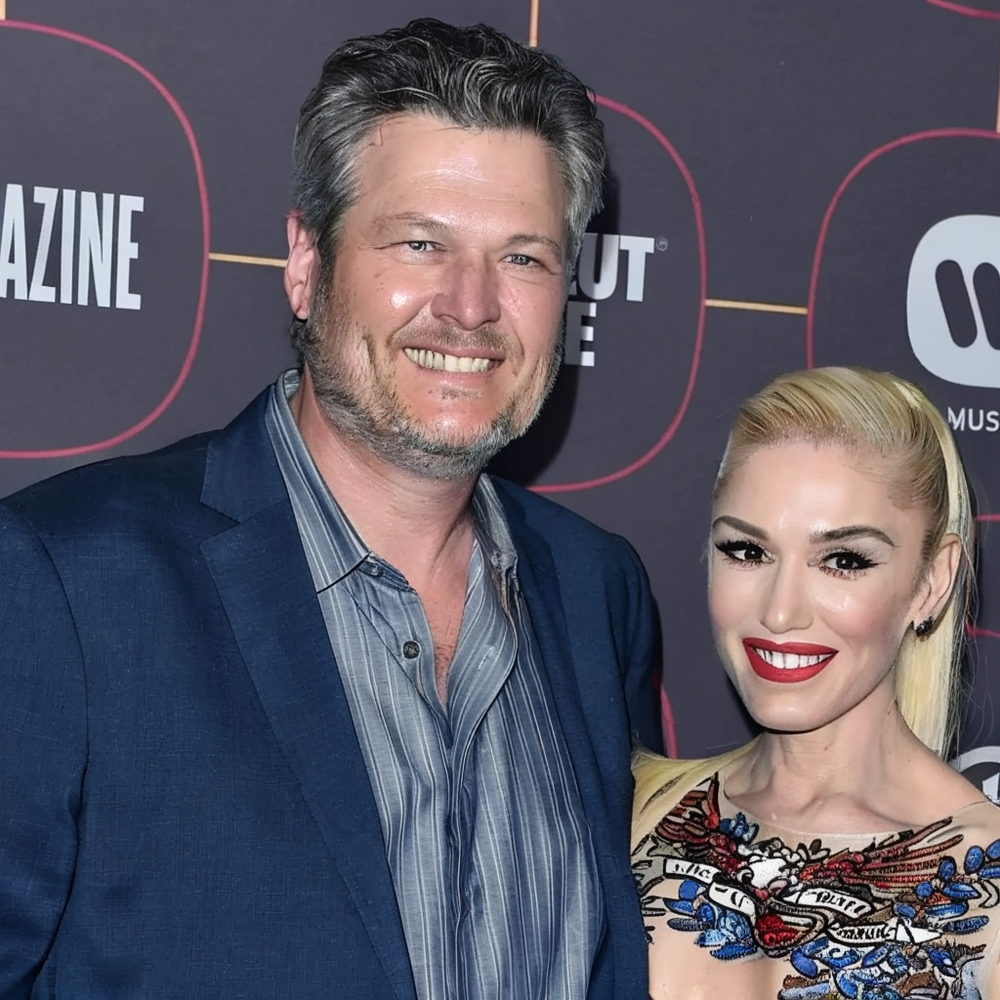}
\end{subfigure}
\begin{subfigure}{0.495\linewidth}
\centering
\includegraphics[width=\linewidth]{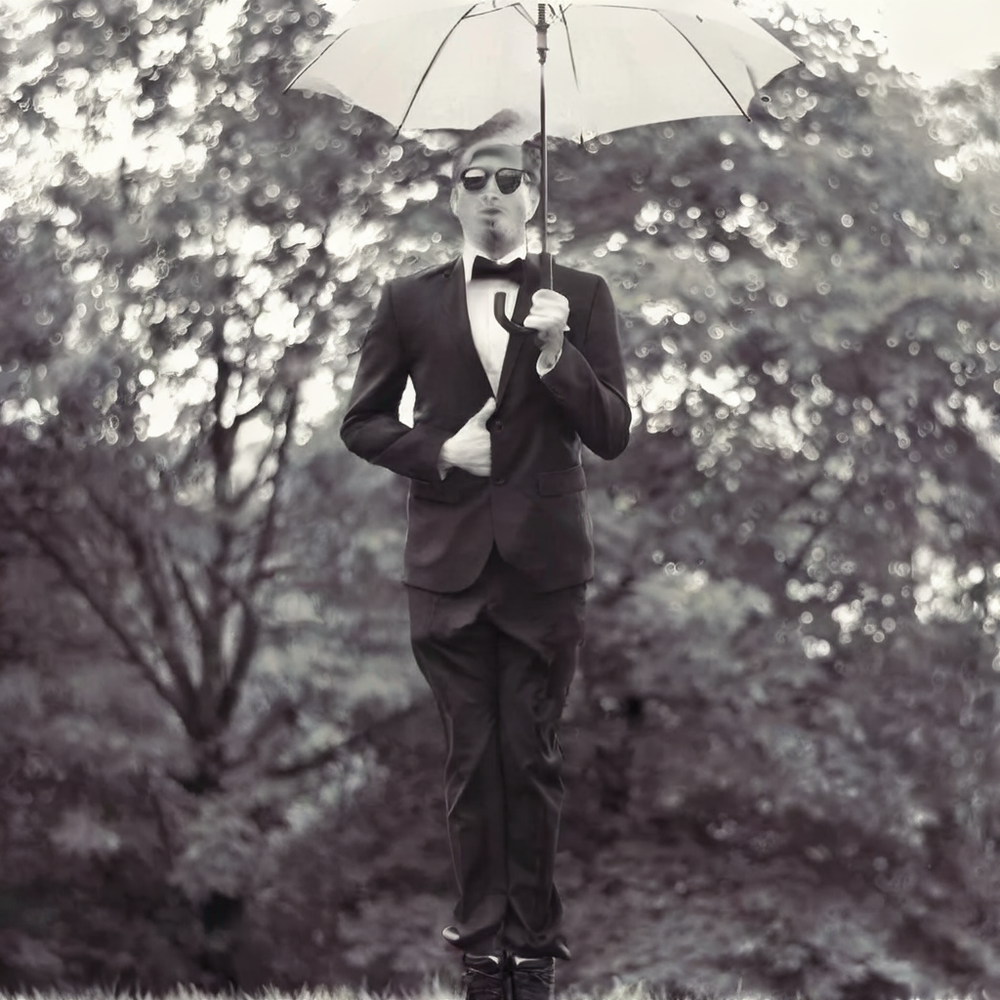}
\end{subfigure}
\hfill
\begin{subfigure}{0.495\linewidth}
\centering
\includegraphics[width=\linewidth]{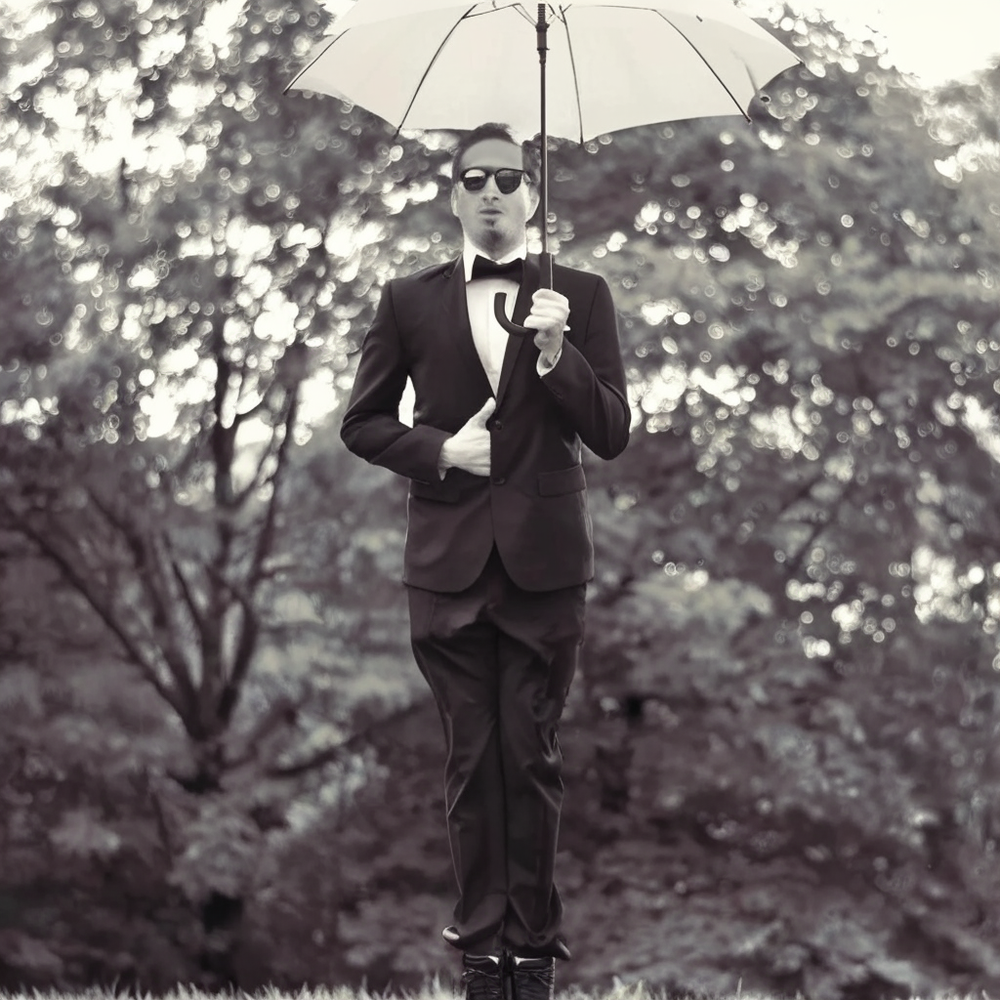}
\end{subfigure}
\caption{Visual comparison of $1024{\times}1024$ image reconstruction between an original RQ-VAE (\textbf{left}) and the one with a fine-tuned decoder (\textbf{right}).}
\label{fig:vae_visual_comparison}
\end{figure}

\end{document}